\documentclass[11pt]{article}

\usepackage[utf8]{inputenc} \usepackage[T1]{fontenc}    \usepackage{url}            \usepackage{booktabs}       \usepackage{amsfonts}       \usepackage{amsmath}
\usepackage{amssymb}
\usepackage{amsthm}
\usepackage{bbm}
\usepackage{nicefrac}       \usepackage{microtype}      \usepackage{xcolor}         \usepackage{graphicx}
\usepackage{caption}
\usepackage{tikz,tikz-3dplot}
\usetikzlibrary{positioning}
\usepackage{array}
\usepackage{xspace}
\usepackage{xurl}
\usepackage{tabularx} 
\usepackage{makecell}
\usepackage{pdflscape}  \usepackage{placeins}
\usepackage{url,ifthen}
\usepackage[shortlabels]{enumitem}
\usepackage{diagbox}
\usepackage{srcltx}
\usepackage{dsfont}
\usepackage{multirow}
\usepackage{boxedminipage}
\usepackage[margin=1.15in]{geometry}

\usepackage{color}
\usepackage{colortbl}
\usepackage{setspace}
\usepackage{pgfplots}
\pgfplotsset{compat=1.12}
\usepackage{algorithm,algorithmic}
\usepackage[algo2e]{algorithm2e}
\usepackage{longtable}

\usepackage[pdftex]{hyperref}
\hypersetup{
    unicode=false,              pdftoolbar=true,            pdfmenubar=true,            pdffitwindow=false,          pdfnewwindow=true,          colorlinks=true,           linkcolor=DarkBlue,              citecolor=DarkGreen,            filecolor=DarkRed,          urlcolor=DarkBlue,              pdftitle={},
    pdfauthor={},
}

\usepackage{microtype}

\renewcommand{\sfdefault}{phv}

\usepackage{listings}
\lstdefinestyle{mystyle}{
    commentstyle=\color{DarkBlue},
    keywordstyle=\color{DarkRed},
    numberstyle=\tiny\color{Gray},
    stringstyle=\color{DarkGreen},
    basicstyle=\footnotesize,
    breakatwhitespace=false,         
    breaklines=true,                 
    captionpos=b,                    
    keepspaces=true,                 
    numbers=left,                    
    numbersep=5pt,                  
    showspaces=false,                
    showstringspaces=false,
    showtabs=false,                  
    tabsize=2
}
\definecolor{DarkGreen}{rgb}{0.1,0.5,0.1}
\definecolor{DarkRed}{rgb}{0.6,0.05,0.05}
\definecolor{DarkBlue}{rgb}{0.1,0.1,0.5}
\definecolor{Gray}{rgb}{0.2,0.2,0.2}

\definecolor{c1}{RGB}{38, 70, 83}
\definecolor{c2}{RGB}{42, 157, 143}
\definecolor{c3}{RGB}{233, 196, 106}
\definecolor{c5}{RGB}{231, 111, 81}
\definecolor{c4}{RGB}{244, 162, 97}
\definecolor{lightgray}{rgb}{0.97,0.97,0.97} 

\NewDocumentCommand{\blue}{ o }{	\textcolor{blue}{#1}}
\usepackage[normalem]{ulem}

\def\draft{0}

\def\submit{1}

\ifnum\draft=1     
\else
    
\fi

\ifnum\submit=1
\newcommand{\forsubmit}[1]{#1}
\newcommand{\forreals}[1]{}
\else
\newcommand{\forreals}[1]{#1}
\newcommand{\forsubmit}[1]{}
\fi

\newcommand{\remove}[1]{}

\newtheorem*{definition*}{Definition}

\theoremstyle{definition}

\newtheoremstyle{example_contd}
{\topsep} {\topsep}{}{}{\bfseries}{.}{1em}{\thmname{#1} \thmnumber{ #2}\thmnote{#3} (continued)}
\theoremstyle{example_contd}

\usepackage{subcaption}

\usepackage{amsmath,amsfonts,bm}

\def\eqref#1{equation~\ref{#1}}

\def\1{\bm{1}}

\DeclareMathAlphabet{\mathsfit}{\encodingdefault}{\sfdefault}{m}{sl}
\SetMathAlphabet{\mathsfit}{bold}{\encodingdefault}{\sfdefault}{bx}{n}

\usepackage[round]{natbib}      \usepackage{authblk}            \usepackage{soul}               
\title{Does Reasoning Make Language Models Agree?\thanks{Corresponding author. Email: \url{mgorecki@tuebingen.mpg.de}}}
\author[1,2]{Mila Gorecki$^*$}
\author[1,2]{Moritz Hardt}
\affil[1]{Max Planck Institute for Intelligent Systems, T\"ubingen}
\affil[2]{T\"ubingen AI Center}

\graphicspath{{./}}

\date{}

\begin{document}
\bibliographystyle{plainnat}
\maketitle

\begin{abstract}
Language models face two opposing concerns:  \emph{monoculture}, in which models err in correlated ways, and \emph{multiplicity}, in which equally accurate models make conflicting decisions. The rise of reasoning models has complicated this debate: On the one hand, reasoning may increase model agreement if models anchor on the same salient features or follow similar reasoning paths. On the other hand, stochastic reasoning traces may introduce additional branch points leading to more diverse predictions. 
We resolve this tension by distinguishing reasoning as an intervention from reasoning as an observed behavior. 
We compare agreement under controlled changes in reasoning effort with agreement across observed reasoning lengths, measuring both against the agreement expected from model accuracy. Across nine models from two developers and three classification tasks, enabling reasoning increases agreement between models evaluated on the same individuals. Yet the individuals who elicit longer reasoning traces receive less consistent predictions across models. Both patterns persist after accounting for differences in accuracy. Asking models to reason more thus promotes agreement, while observing them reason longer signals disagreement.
\end{abstract}

\section{Introduction}\label{sec:intro}
\FloatBarrier
\suppressfloats[t]

When different models assess the same individual in consequential settings, two seemingly opposing concerns arise. Under \emph{algorithmic monoculture}, decision makers rely on the same or similar models, potentially leaving individuals unable to escape erroneous assessments by switching models \citep{kleinberg2021AlgorithmicMonocultureSocial,creel2021AlgorithmicLeviathanArbitrariness,bommasani2022PickingSamePerson}. Under \emph{model multiplicity}, similarly accurate models make conflicting predictions, so model choice can determine an individual's outcome \citep{marx2020PredictiveMultiplicityClassification,black2022ModelMultiplicityOpportunities}.

Prior work on language models finds correlated errors across models \citep{goel2025GreatModelsThink,kim2025CorrelatedErrorsLarge}. On predictions about individuals, similarly accurate language models agree more often than an independent-error baseline would suggest, but remain well below complete agreement, an extreme form of monoculture \citep{gorecki2025MonocultureMultiplicityWhich}. That study examines base and instruction-tuned models, leaving open how reasoning changes this picture. \emph{Does more reasoning push predictions towards greater monoculture or greater multiplicity?}

\begin{figure}[t]
    \centering
    \includegraphics[width=\linewidth]{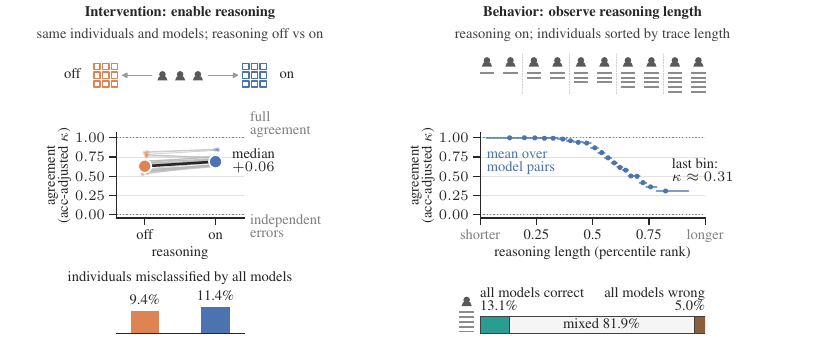}
    \caption{Reasoning and model agreement on ACSIncome. \textit{Left:} enabling reasoning increases accuracy-adjusted agreement between models evaluated on the same individuals. \textit{Right:} with reasoning enabled, individuals eliciting longer traces receive less consistent predictions across models. The panels distinguish the effect of changing reasoning settings from the association between observed trace length and agreement.}
    \label{fig:fig1}
\end{figure}

Reasoning introduces additional computation before a model commits to an answer, with demonstrated gains on tasks such as mathematics and code \citep{openai2024OpenAIO1System,deepseek-ai2025DeepSeekR1IncentivizingReasoning}. Its implications for agreement are less clear. Models could converge by attending to similar features or following similar reasoning patterns, consistent with evidence of correlated errors across increasingly capable models \citep{goel2025GreatModelsThink,kim2025CorrelatedErrorsLarge}. Alternatively, stochastic intermediate generations could introduce variability. Questions eliciting longer traces have been found to receive less consistent answers from a single model \citep{hagele2026thehotmess}. These findings make both greater agreement and greater disagreement plausible, but concern different forms of consistency and do not settle what reasoning does to agreement between models.

We resolve this apparent tension by distinguishing reasoning as a treatment from reasoning length as observed behavior. Under \emph{reasoning as an intervention}, we vary reasoning settings while holding the model and prompt fixed. Under \emph{reasoning as behavior}, we compare individuals who elicit different trace lengths within a fixed setting. The first analysis measures the effect of enabling reasoning; the second identifies where models disagree.
This distinction yields a clear empirical picture (\autoref{fig:fig1}): asking models to reason longer promotes agreement, while observing them reason longer about an individual signals less consistent predictions across models. Both patterns persist after accounting for accuracy.

\subsection{Our Contributions}
In this work, we examine how reasoning shapes inter-model agreement and individuals' access to a correct prediction through model switching. Our analysis distinguishes intervention effects from observed associations by comparing the same models with reasoning enabled and disabled, and separately examining how agreement varies with realized trace length within fixed reasoning settings. In both analyses, we measure pairwise agreement and the number of models predicting each individual correctly, comparing both with what independent errors at the models' observed accuracies would imply. This distinguishes agreement on correct predictions from shared errors that persist across model choices and shows which individuals receive a correct prediction from some, all, or none of the evaluated models.

Across nine models, two developers, and three prediction tasks, we find opposing patterns that persist after accounting for accuracy: enabling reasoning increases agreement (\autoref{fig:agree-effort}), while individuals eliciting longer traces receive less consistent predictions (\autoref{fig:agree-length}). Enabling reasoning also concentrates individual outcomes at both unanimous correctness and unanimous error (\autoref{fig:effort-recourse}), raising the share of individuals misclassified by all nine models by $2.0$-$2.4$ percentage points. Individuals who elicit longer traces more often receive a mix of correct and incorrect predictions across models. Yet unanimous errors persist: as in the comparison between reasoning settings, misclassification by all models is more frequent than the independent-error baseline predicts (\autoref{fig:length-recourse}). Thus, greater agreement does not guarantee access to a correct prediction, and greater disagreement does not eliminate shared errors.

\subsection{Related Work}

\paragraph{Monoculture and Multiplicity.}
Work on algorithmic monoculture examines shared errors and exclusion when decision makers rely on the same or similar models \citep{bommasani2022PickingSamePerson,bommasani2022OpportunitiesRisksFoundation}, as well as aggregate welfare losses \citep{kleinberg2021AlgorithmicMonocultureSocial, kleinberg2026PriceAnarchyAlgorithmic}. Empirical studies find shared failures across commercial models \citep{chen2022HAPILargescaleLongitudinal,toups2024EcosystemlevelAnalysisDeployed} and repeated rejection of applicants when assessed by a shared hiring vendor \citep{bommasani2026AlgorithmicMonoculturesHiring}.
Work on model multiplicity studies how conflicting individual predictions, explanations, or other outcomes can arise among models with similar predictive performance \citep{breiman2001StatisticalModelingTwo,marx2020PredictiveMultiplicityClassification,black2022ModelMultiplicityOpportunities,damour2020UnderspecificationPresentsChallenges,ganesh2025SystemizingMultiplicityCurious}.
The normative implications of consistency and divergence across model outputs depend on the decision context \citep{creel2021AlgorithmicLeviathanArbitrariness,jain2024AlgorithmicPluralismStructural,hedden2026AlgorithmicMonocultureIts,gur-arieh2025ConsistentlyArbitraryArbitrarily,gur-arieh2026AmbiguityCollapseLLMs,jain2026TaskDependentEvaluationLLM}. Our contribution is empirical: We study how reasoning affects agreement and shared errors.

Prior work measures agreement relative to independent errors at observed accuracies, both between model pairs \citep{mania2019ModelSimilarityMitigates,goel2025GreatModelsThink,kim2025CorrelatedErrorsLarge} and across model sets \citep{toups2024EcosystemlevelAnalysisDeployed}.
Recent work suggests that conclusions about excess agreement may depend on the choice of baseline \citep{jo2026SubjectivityMonoculture}. For this work, we follow prior work and adopt an accuracy-based baseline.

Language models exhibit homogeneity in open-ended generation \citep{wu2024GenerativeMonocultureLarge,jiang2025ArtificialHivemindOpenEnded} and correlated errors on benchmarks and simulated hiring tasks \citep{goel2025GreatModelsThink,kim2025CorrelatedErrorsLarge}. Most directly related, \citet{gorecki2025MonocultureMultiplicityWhich} study pairwise agreement and individual recourse for base and instruction-tuned models on tabular prediction tasks. They find agreement above an independent-error baseline but below strict monoculture. We extend this analysis to reasoning models, studying reasoning settings and variation in observed trace length.

\paragraph{Reasoning.}
Chain-of-thought prompting \citep{wei2022ChainofthoughtPromptingElicits,kojima2022LargeLanguageModels} and reinforcement learning for reasoning \citep{openai2024OpenAIO1System,deepseek-ai2025DeepSeekR1IncentivizingReasoning} have significantly driven performance on tasks with verifiable answers.
Studies differ in how they control or measure reasoning: Reasoning can be varied through provider-exposed effort settings or token budgets \citep{mei2026ReasoningUncertaintyReasoning,wani2026ReasoningTaxToken}\ or prompt instructions \citep{renze2024BenefitsConciseChaina,lee2025HowWellLLMs,nayab2026ConciseThoughtsImpact,han2025TokenBudgetAwareLLMReasoning,gema2025InverseScalingTestTime}. 
Increasing test-time compute through forced continuation \citep{muennighoff2025S1SimpleTesttime} or by sampling and aggregating multiple reasoning paths within-model \citep{wang2023SELFCONSISTENCYIMPROVESCHAIN,snell2024ScalingLLMTestTime} or even across models \citep{liu2026LLMsJuryCrossModel} can improve task accuracy. 
While these studies examine gains from increasing test-time compute, a complementary line of work relates observed reasoning length to performance.
 Longer traces correlate with lower accuracy on competition math, even after controlling for problem difficulty \citep{ballon2026RelationshipReasoningPerformance}, and lower model confidence \citep{devic2025TraceLengthSimple}. Longer reasoning can also accompany systematic failure: on tasks designed to expose overthinking, extended reasoning amplified distraction, reliance on spurious features, or failures to track constraints \citep{gema2025InverseScalingTestTime}. 
Taken together, these studies complicate the relationship between reasoning and performance. \citet{hagele2026thehotmess} examine this tension through consistency within individual models: within a fixed reasoning budget, longer observed traces are associated with greater within-model error-incoherence, while increasing reasoning budgets modestly reduces it for most tested models. Their error-incoherence metric measures the fraction of error attributable to variance across samples. The coexistence of these findings underscores the ambiguity of claims about ``more reasoning''. We make the treatment-behavior distinction central to our study of how reasoning shapes agreement \emph{between} models.

Trace length need not measure computational difficulty \citep{stechly2025SemanticsUnreasonableEffectiveness}, and stated reasoning need not faithfully explain a prediction \citep{turpin2023LanguageModelsDont}. We therefore treat length as an observed marker\ and examine its association with agreement.

Another strand of work connects reasoning models to the concerns of monoculture and multiplicity more directly through output diversity and shared errors. Post-training on reasoning traces can reduce output diversity, which is not restored by suppressing chain-of-thought at inference \citep{karouzos2026WhereDoesOutput}. Across reasoning models, \citet{liu2026LLMsJuryCrossModel} find lower error correlation between model families than within them, alongside persistent shared errors on some tasks. These findings concern verifiable tasks or open-ended generation. We examine predictions about individuals, connecting inter-model agreement to their access to a correct prediction.

\section{Measuring Agreement Beyond Accuracy}\label{sec:metrics}
\paragraph{Preliminaries.} We consider $n$ individuals with features $x_i$ and binary labels $y_i\in\{0,1\}$, 
all evaluated by $M$ models $h_1,\dots,h_M$. A model $h_m$ paired with a reasoning setting $r$ (e.g., reasoning disabled) defines a \emph{configuration}, which we evaluate in multiple independent resamples indexed by $s$. For each individual $i$, model $h_m$ generates a risk score under reasoning setting $r$ in resample $s$. We threshold this score to obtain a binary prediction $\hat y_{i,m,r,s}\in\{0,1\}$ (Section~\ref{sec:exp_setup}). For the definitions below, we hold $r$ and $s$ fixed and write $\hat y_{im}$ for notational simplicity.\ Let $a_m=\frac1n\sum_i \mathbbm{1}[\hat y_{im}=y_i]$ denote the accuracy of model $h_m$\ in this setting and resample. All quantities  are computed on the individuals included in the respective analysis. For subsets such as reasoning-length bins, we recompute accuracies and the corresponding baselines using only those individuals.

Based on the model predictions, we ask two questions. Do models err on the same individuals, leaving them without recourse (\emph{monoculture})? Or do equally accurate models assign them conflicting predictions (\emph{multiplicity})? We address them at two levels: between pairs of models, and for each individual across the whole model set.

\paragraph{Pairwise Agreement.}
The observed agreement between models $h_j$ and $h_k$ is the fraction of individuals on which they make the same prediction,
\begin{equation}\textstyle\label{eq:po}
    p_o(h_j,h_k)=\frac1n \sum_{i=1}^n \mathbbm{1}[\hat y_{ij}=\hat y_{ik}].
\end{equation}
For readability, we drop $h$ and write $p_o(j,k)$ instead of $p_o(h_j,h_k)$, and likewise for other pairwise metrics introduced below. Models can agree on both correct and incorrect predictions. Observed agreement is hard to interpret on its own, because it is constrained by accuracy: In the binary setting, two models with accuracies $a_j$ and $a_k$ agree on at least $|a_j+a_k-1|$ and at most $1-|a_j-a_k|$ of individuals (Fréchet bounds). So, two models with $0.8$ accuracy each must  agree on at least $0.6$ of individuals and agree on all of them, when all errors align.

\paragraph{Accuracy-based null model.} We compare observed agreement with an \textbf{accuracy-based null model $p_\mathrm{base}$} in which models place errors independently and uniformly across individuals while preserving their accuracies:
\begin{equation}\textstyle\label{eq:pbase}
    p_\mathrm{base}(j,k)=a_j \cdot a_k + (1-a_j)\cdot (1-a_k).
\end{equation}
This baseline is widely used in the literature \citep{geirhos2020AccuracyQuantifyingTrialbytrial,goel2025GreatModelsThink,gorecki2025MonocultureMultiplicityWhich}. Following \citet{gorecki2025MonocultureMultiplicityWhich}, we interpret it as strong predictive multiplicity.
Under this baseline, two models with $0.8$ accuracy agree on $0.68$ of individuals by chance. The \emph{agreement beyond chance}, $\mathrm{ABC}(j,k)=p_o(j,k)-p_\mathrm{base}(j,k)$, measures excess agreement relative to this baseline.

Throughout this work, we report the \textbf{\emph{accuracy-adjusted} $\mathbf{\kappa}$}, which scales excess by perfect agreement:\begin{equation}\textstyle\label{eq:kappa}
    \kappa(h_j,h_k) = \frac{p_o - p_\mathrm{base}}{1 - p_\mathrm{base}}.
\end{equation}
Unlike Cohen's $\kappa$, which uses positive prediction rates, the accuracy-adjusted $\kappa$ derives expected agreement from model accuracies. It thus asks whether two models err on the same individuals more often than their accuracies imply. $\kappa=0$ means agreement matches the baseline, $\kappa>0$ that they are more aligned than the baseline, and $\kappa=1$ that they agree on every individual. When $p_o=p_\mathrm{base}=1$, the ratio is undefined and we set
$\kappa=1$ by convention in both the effort and length analyses. For binary predictions, $\kappa$ equals the error consistency of \citet{geirhos2020AccuracyQuantifyingTrialbytrial}, which is defined on agreement in correctness rather than in predictions.

Model accuracies determine both the normalization factor of \mbox{\eqref{eq:kappa}} and the attainable range of agreement
, a known difficulty of chance-corrected agreement \mbox{\citep{feinstein1990HighAgreementLow,byrt1993BiasPrevalenceKappa}}. We therefore  report all components $p_o$, $p_\mathrm{base}$ and ABC separately in the Appendix. Throughout the paper, for each pair of distinct models, we compute $p_o$ and $p_\mathrm{base}$ for every combination of their resamples, average both components across these combinations, and form $\kappa$ from these averages. This avoids dividing by potentially small denominators $1-p_\mathrm{base}$ separately for each combination. We use this procedure in both analyses.

\paragraph{Individual-level outcomes and recourse.}
Pairwise agreement does not show how errors are distributed across individuals. Following \mbox{\citet[Def.~5]{gorecki2025MonocultureMultiplicityWhich}}, we therefore consider \emph{recourse within the model set}: an individual's ability to obtain a favorable outcome by switching models. We operationalize a favorable outcome as a correct prediction and additionally report results conditioned on the true label. For each individual $i$, let
\begin{equation}\textstyle
    m_i = \sum_{j=1}^M \mathbbm{1}[\hat y_{ij} = y_i] \in \{0,\dots,M\}
\end{equation}
denote the number of models that predict their label correctly. The \emph{recourse level} $\frac{m_i}{M}$ is the fraction of models that provide a correct prediction. Higher values indicate more opportunities to obtain a correct decision by switching models. At $m_i=0$ every model misclassifies the individual: there is no recourse, a form of \emph{systemic failure} associated with monoculture \citep{bommasani2022PickingSamePerson}. Individuals with $m_i=M$ are classified correctly by every model and have full recourse. Intermediate values indicate that the outcome depends on which model is used, linking recourse to predictive multiplicity.

\paragraph{Independent-error baseline.}
The same independent-error assumption yields a baseline for the recourse distribution \citep{bommasani2022PickingSamePerson,toups2024EcosystemlevelAnalysisDeployed,gorecki2025MonocultureMultiplicityWhich}. If each model $h_j$ is correct independently with probability $a_j$, then $m_i$ follows a Poisson-binomial distribution:
\begin{equation}\label{eq:poisson-binomial}
    \Pr[m_i = k]
     = \sum_{S\subseteq[M],\,|S|=k}\;
        \prod_{j\in S} a_j 
        \prod_{j\in [M]\setminus S}(1-a_j),
        \qquad k\in\{0,\dots,M\},
\end{equation}
where $[M]=\{1,\dots,M\}$ and subset $S$ indexes the models predicting correctly. For $M=2$, the probability of a unanimous outcome, 
$\Pr_{\mathrm{base}}[m_i=0]+\Pr_{\mathrm{base}}[m_i=2]=p_\mathrm{base}(1,2)$
recovers the pairwise agreement baseline
 in \eqref{eq:pbase}, so this baseline generalizes the pairwise one.
We report the distribution of $m_i$ across individuals and its \emph{excess} over the baseline: the observed fraction at each $k$ minus its baseline probability (\eqref{eq:poisson-binomial}). Positive excess at $k=0$ or $k=M$ indicates that no recourse or full recourse, respectively, occurs more often than the baseline would imply. Because unanimity is easier to reach with fewer models, we compare groups of equal size $M$ evaluated on the same individuals. For each resample $s$, we use one prediction per model and individual\ and compute both the observed distribution and its baseline. We then average each distribution across resamples.

Both baselines assume that each model is equally likely to be correct on every individual. Null models that also account for item difficulty can lead to different conclusions \citep{jo2026SubjectivityMonoculture}. Our comparisons across reasoning settings and length bins thus remain relative to this accuracy-based baseline.

\section{Experimental Design}\label{sec:exp_setup}

\paragraph{Overview.}
We evaluate nine language models from two developers on three binary prediction tasks: ACSIncome, ACSEmployment, and SIPP. We assess pairwise agreement and individual-level recourse under changes in reasoning settings (\emph{reasoning as an intervention}) and across individuals grouped by realized reasoning length (\emph{reasoning as behavior}).

\paragraph{Models and prompting.}
We evaluate nine reasoning language models that support disabling reasoning for matched comparisons: six from Alibaba Cloud (Qwen 3 4B, 14B, 32B and Qwen 3.5 4B, 9B and 27B) and three from OpenAI (GPT 5.1, 5.2 and 5.4). Appendix~\ref{appendix:models} provides details. We evaluate OpenAI models at effort levels \texttt{none}, \texttt{low}, \texttt{medium}, and \texttt{high}, and Alibaba models with reasoning enabled and disabled.

We use the default numeric QA template in \texttt{folktexts}~\citep{cruz2024EvaluatingLanguageModels} to convert tabular features of individuals into standardized, zero-shot prompts asking for the probability of the positive class. We adapt the package to support reasoning models and add an answer-formatting instruction to the system prompt. Answer probabilities are extracted using regular expressions. All models receive the same zero-shot prompt, without additional instructions how long or extensively to reason. Appendix~\ref{appendix:inference-protocol} provides details on the prompting and inference protocol. We convert extracted probabilities into class predictions using a decision threshold $t$ fitted separately for each model, reasoning setting, and resample to maximize balanced accuracy on a validation set of $2000$ individuals, following \citet{cruz2024EvaluatingLanguageModels} and \citet{gorecki2025MonocultureMultiplicityWhich}.

\paragraph{Reasoning settings and measurement.}
Under the intervention lens, we compare \texttt{off} versus \texttt{on} for Alibaba models and \texttt{none} versus \texttt{high} for OpenAI models. Additional results for intermediate effort levels of OpenAI models (\texttt{low} and \texttt{medium}) are reported in the appendix. These comparisons capture the effect of enabling reasoning within each model, but we note that effort levels are not directly comparable across providers.
Realized reasoning length varies across individuals within each setting, motivating the behavioral analysis in Section~\ref{sec:agree-vs-length}.
Under the behavioral lens, we count tokens within \texttt{<think></think>} blocks emitted by Alibaba models and use the reasoning-token counts reported in the API response metadata for OpenAI models. We record the available reasoning traces or summaries, token counts, responses, and extracted answers. Appendix~\ref{appendix:inference-protocol} provides details on answer extraction and reasoning-length measurement.

\begin{figure}[tbp]
    \centering
    \includegraphics[width=\linewidth]{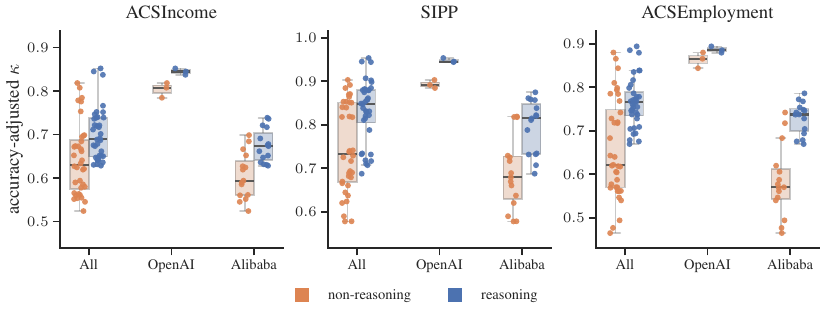}
    \caption{Accuracy-adjusted agreement $\kappa$ between distinct model pairs in the same reasoning condition: disabled (orange) or enabled at the highest evaluated effort (blue). Each dot represents one pair. Observed and baseline agreement are averaged over five matched resamples before computing $\kappa$.}     \label{fig:agree-effort}
\end{figure}

\paragraph{Binary prediction tasks.}
We evaluate three binary classification tasks from two data sources. ACSIncome and ACSEmployment predict whether an individual's annual income exceeds \$50,000 and whether they are employed, respectively. Both tasks are based on the U.S. Census Bureau's American Community Survey (ACS) Public Use Microdata Sample \citep[PUMS,][]{u.s.censusbureau2018AmericanCommunitySurvey}, with task definitions from \citet{ding2022RetiringAdultNew} and natural-text feature mappings from \texttt{folktexts} \citep{cruz2024EvaluatingLanguageModels}. We select these tasks because prior work found that base and instruction-tuned language models underperform an XGBoost baseline, leaving room for potential gains from reasoning \citep{gorecki2025MonocultureMultiplicityWhich}.
SIPP uses the longitudinal Survey of Income and Program Participation \citep{u.s.censusbureau2014SurveyIncomeProgram} to predict whether an individual's income is significantly above the Official Poverty Measure (OPM). We adapt its column-to-text mappings from \citet{gorecki2025MonocultureMultiplicityWhich}. See Appendix~\ref{appendix:tasks} for a detailed description of all tasks.

To limit computational costs, we randomly sample 8323 instances from ACSIncome ($5\%$ of the full dataset) and 5000 from ACSEmployment ($1.5\%$), and use all 3972 SIPP test records ($10\%$ of the full dataset), excluding the $2000$ validation individuals used for threshold fitting. All models and evaluated reasoning configurations receive the same inputs. We generate five resamples per configuration and input to capture generation stochasticity. For each pair of distinct models, agreement is aggregated over all $5\times5$ combinations of runs as described in Section~\ref{sec:metrics}.

Our analysis proceeds in two stages, corresponding to the intervention and behavioral lenses. We first examine how enabling reasoning changes agreement between models and individuals' opportunities to obtain a correct prediction by switching models. We then hold reasoning settings fixed and examine how these outcomes vary across individuals who elicit shorter or longer reasoning traces.

\section{Reasoning as Intervention: Agreement vs.\ Reasoning Effort}\label{sec:agree-vs-effort}
We first ask whether enabling reasoning increases model agreement. Reasoning settings provide a practical way to vary inference-time computation: provider documentation describes higher effort as a trade-off between response quality, latency, and token usage \mbox{\citep{comanici2025Gemini25Pushing,openai2025reasoningeffort,anthropic2026effortdocs}}. We compare the same models and prompts with reasoning disabled and enabled at the highest evaluated setting. Unless stated otherwise, numerical ranges in the following results span the three tasks.

Aggregate accuracy varies considerably between models on each task (\autoref{fig:appendix-general-accuracy-all-tasks}) and enabling reasoning does not consistently improve aggregate accuracy. Nevertheless, most configurations fall within an empirical Rashomon set with $\epsilon=0.05$ \citep{marx2020PredictiveMultiplicityClassification,semenova2022ExistenceSimplerMachine}, defined in Appendix~\ref{appendix:general-performance}. Only a few configurations, in particular the smaller, mostly 4B-parameter Qwen models with reasoning disabled, fall outside of it. We keep them to preserve the matched setting, noting that this yields broader Rashomon sets with $\epsilon \in [0.06,0.10]$ across tasks, based on the worst-performing resample.

\paragraph{Pairwise agreement.}
Across all three tasks, enabling reasoning (\texttt{on}~/~\texttt{high}) shifts accuracy-adjusted pairwise agreement towards higher $\kappa$ values (\autoref{fig:agree-effort}). Because we account for accuracy, this increase cannot be explained by increased model performance alone. Across all model pairs, the median $\kappa$ increases by $0.06$--$0.14$. The increase is largest for pairs in which both models were developed by Alibaba Cloud ($0.08$--$0.17$, \texttt{on} vs \texttt{off}). OpenAI model pairs, which already agree strongly with reasoning disabled (median $\kappa$ $0.81$--$0.89$), gain less ($0.02$--$0.05$, \texttt{none} vs \texttt{high}). Across all pairs, $\kappa$ values are also less dispersed with reasoning enabled on every task.
The positive $\kappa$ values throughout indicate that observed agreement across model pairs markedly exceeds the baseline agreement expected if models made random errors at their respective accuracy levels. Note that the increase in agreement is not accompanied by systematically more extreme risk-score distributions (\autoref{fig:appendix-general-risk-scores-all-tasks}).
These patterns are also reflected in raw observed agreement. All model pairs in the non-reasoning condition agree on at least $80\%$ of individuals, whereas reasoning model pairs agree on at least $86\%$ of individuals. Decomposing pairwise agreement by prediction correctness (\autoref{fig:appendix-effort-pairwise-agreement-by-correctness-all-tasks}) shows that for Alibaba model pairs reasoning increases agreement on both correct and incorrect predictions. Results for OpenAI model pairs vary by task: on ACSIncome, agreement shifts towards incorrect predictions; on SIPP, agreement on both correct and incorrect predictions increases; on ACSEmployment, agreement on correct predictions increases while agreement on incorrect predictions decreases.
\autoref{fig:appendix-effort-kappa-components-all-tasks} reports all components of the accuracy-adjusted $\kappa$, namely observed agreement, baseline agreement and their difference. 
Among OpenAI models, which support intermediate effort levels, $\kappa$ increases with effort on SIPP and ACSEmployment and peaks at \texttt{medium} on ACSIncome (\autoref{fig:appendix-effort-kappa-by-level-all-tasks}).

\begin{figure}[tbp]
    \centering
    \includegraphics[width=\textwidth]{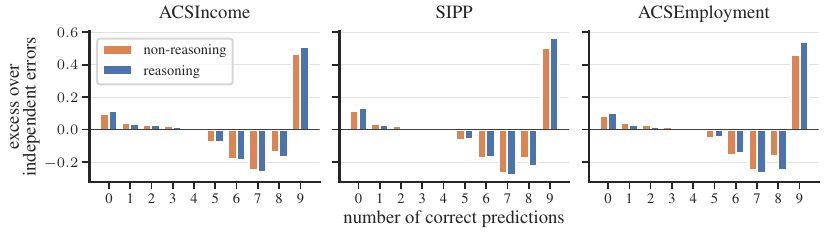}
    \caption{Excess over the independent-error baseline by number of correct models ($m$ out of nine), with reasoning disabled (orange) or enabled (blue). Excess is the observed share minus the share expected from model accuracies under independent errors; observed distributions appear in \autoref{fig:appendix-effort-recourse-distribution}.}
    \label{fig:effort-recourse}
\end{figure}

\paragraph{Individual-level agreement and shared errors.} 
To determine whether greater agreement reflects correct predictions or shared errors, we examine individual recourse levels across all nine models in each reasoning condition.
With reasoning enabled, we observe larger fractions of individuals receiving unanimous predictions, either correct (full recourse, $m=9$) or incorrect (no recourse, $m=0$), while few individuals have recourse levels near $0.5$, where disagreement among models is highest (\autoref{fig:appendix-effort-recourse-distribution}).
Averaged across resamples, the share of individuals without recourse rises from $9.44\%$ to $11.41\%$ on ACSIncome, from $11.29\%$ to $13.62\%$ on SIPP, and from $8.13\%$ to $10.52\%$ on ACSEmployment. These increases amount to approximately $2.0$--$2.4$ percentage points, or $21$--$29\%$ relative to reasoning disabled. This increase holds across all resamples and tasks. 
To account for differences in accuracy, \autoref{fig:effort-recourse} shows the excess of these shares over the independent-error baseline.
 Both no and full recourse occur more often than the baseline predicts, and  excess grows when reasoning is enabled. The pattern persists when conditioning on the true label (\autoref{fig:appendix-effort-recourse-by-label-all-tasks}), so the concentration at the extremes is not driven by one  class.
In the high-recourse regime ($m\in\{5,\dots,8\}$), however, where the majority of models predicts the correct label, the reasoning condition shows a more negative deviation from the baseline than the non-reasoning condition does. Thus, greater agreement concentrates outcomes at both full and no recourse, rather than uniformly improving access to a correct prediction.
The exact transitions for individuals when moving from non-reasoning to reasoning are shown in \autoref{fig:appendix-effort-recourse-transitions-all-tasks}. Most individuals keep their recourse level when reasoning is enabled (67--76\%); among the others, increases outnumber decreases. The most frequent change on each task is from $8$ to $9$ correct models. Transitions from $1$ to $0$ are also common, leaving some individuals without their sole source of a correct prediction.

\paragraph{Conclusions.} 
Enabling reasoning increases both agreement and the share of individuals misclassified by all nine models on all three tasks. We next examine how agreement varies with the reasoning lengths individuals elicit.

\section{Reasoning as Behavior: Agreement vs.\ Realized Reasoning Length}\label{sec:agree-vs-length}
The preceding section examined how enabling reasoning changes agreement. We now ask: within a fixed reasoning setting, how does agreement vary across individuals who elicit different reasoning lengths? Reasoning lengths vary across individuals for both developers and overlap substantially across OpenAI effort levels (\autoref{fig:appendix-length-token-distributions-by-effort-all-tasks}). We use the same  predictions as in the preceding section, but group individuals by the lengths of the reasoning traces they elicit. Because length is observed rather than controlled, comparisons between these groups measure association, not the effect of extending reasoning.

\begin{figure}[tbp]
     \centering
    \includegraphics[width=\linewidth]{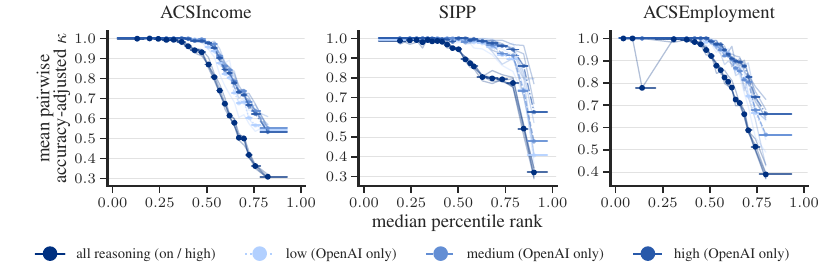}
    \caption{Accuracy-adjusted agreement by reasoning length. Individuals are binned by their median length percentile across nine models and all resamples. Bins contain the same individuals across curves, allowing within-bin comparisons. Dots mark bin medians, horizontal segments span bins, and thin lines show individual resamples. Dark blue: all nine models (Alibaba \texttt{on}, OpenAI \texttt{high}); lighter blues: OpenAI models at each indicated effort.}
    \label{fig:agree-length}
\end{figure}

\paragraph{Pairwise agreement.}
Across all tasks and sets of model configurations shown in \autoref{fig:agree-length}, individuals who elicit relatively longer reasoning traces receive less consistent predictions across model pairs, whereas those who elicit shorter traces experience high pairwise agreement.
We group individuals by their median reasoning-length percentile rank across all resamples of all nine models. We report percentile ranks, which measure relative length within each model, since absolute token counts are not directly comparable across models. Accuracies and the agreement baseline are recomputed within each bin, so $\kappa$ measures agreement beyond what those accuracies imply (Section~\ref{sec:metrics}). Bins in which all models are correct on every individual have $\kappa=1$ by convention (Section~\ref{sec:metrics}). Note that observed agreement also declines (\autoref{fig:appendix-length-agree-observed}), so the drop is not an artifact of the accuracy adjustment.

For individuals up to the median reasoning length, $\kappa\approx1$ in nearly all bins, i.e., models agree in their prediction for almost all of them. Above the median, agreement declines with reasoning length. Among individuals eliciting the longest traces, $\kappa$ across all nine reasoning models falls to approximately $0.31$ (ACSIncome), $0.32$ (SIPP), and $0.39$ (ACSEmployment), still well above the independent-error baseline of $0$.
Within the same bins, agreement tends to be lower across all nine models than among the three OpenAI models at fixed effort levels.The pooled group includes cross-developer pairs, so this difference also reflects which models are compared. 
Sets of models from a single developer might be more aligned because they share much of their training data and post-training procedures.
At the long-reasoning tail, pairwise agreement among OpenAI models also differs by effort level: on SIPP and ACSEmployment, $\kappa$ in the longest bins is lower at \texttt{low} than at \texttt{high} effort, whereas on ACSIncome the three levels end at similar values. This matches the intervention result in Section~\ref{sec:agree-vs-effort}, where agreement among OpenAI models increases with effort level on SIPP and ACSEmployment (\autoref{fig:appendix-effort-kappa-by-level-all-tasks}).

The decline is robust to alternative binning schemes and length measures. It persists when bins are formed separately for each developer (\autoref{fig:appendix-length-agree-by-dev}). When bins are formed separately for each model pair, the decline appears for every model pair on all three tasks (\autoref{fig:appendix-length-agree-pairs-by-dev}). Measuring length in raw tokens or as a per-model z-score instead of percentile ranks yields the same trend (\autoref{fig:appendix-length-agree-by-dev-x-axis}).

Taken together, individuals who elicit longer traces therefore face greater dependence on model choice, a pattern associated with multiplicity. We next examine how this disagreement coexists with shared errors.

\begin{figure}[tbp]
    \centering
    \includegraphics[width=\linewidth]{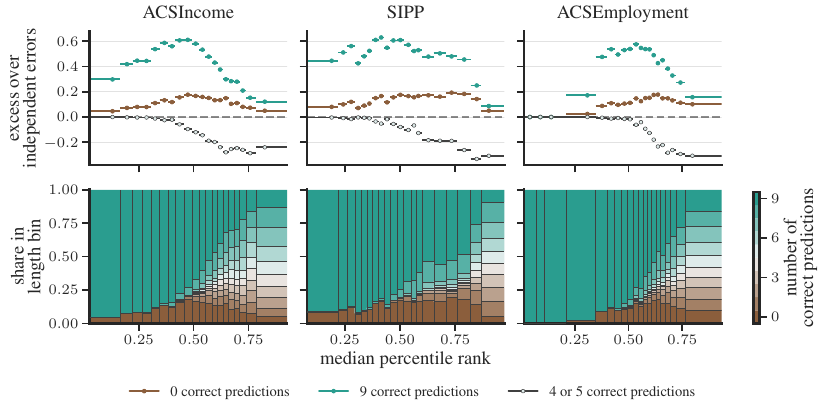}
    \caption{Recourse by reasoning-length bin, as in \autoref{fig:agree-length}, with reasoning enabled for all nine models (Alibaba \texttt{on}, OpenAI \texttt{high}). \textit{Top:} excess over the independent-error baseline at at 0, 9, and 4-5 correct models (no recourse, full recourse, and maximal split). \textit{Bottom:} shares of individuals by number of correct models, from 0 (brown) to 9 (green). Distributions and baselines are averaged across five resamples.}
    \label{fig:length-recourse}
\end{figure}

\paragraph{Individual-level agreement and shared errors.}
To distinguish disagreement from shared errors, we count how many of the nine models predict each individual correctly. Across all three tasks, we find that individuals who elicit short reasoning traces almost always receive the same prediction from all nine models (correct or incorrect), whereas most individuals who elicit long traces receive a mix of correct and incorrect predictions (\autoref{fig:length-recourse}).
Within each length bin, we compare the distribution of correct-prediction counts with its Poisson-binomial baseline, computing both separately for each resample and averaging across resamples.

Below the median reasoning length, nearly all individuals receive either unanimously correct predictions (87--94\%) or unanimously incorrect predictions with no recourse by switching models (5--12\%). Intermediate recourse levels are almost absent (1--3\%). Above the median, the share of individuals with intermediate recourse levels grows steadily, and in the longest bin only about 9--16\% of individuals receive a correct prediction from all nine models (ACSIncome:~13\%~/~SIPP:~9\%~/~ACSEmployment:~16\%). No recourse is most common for individuals at intermediate lengths (18--19\%) and less common in the longest bin (5--10\%). Relative to the independent-error baseline (top row of \autoref{fig:length-recourse}), no and full recourse exceed it in every bin in which models make errors.
The excess at full recourse peaks close to the median length and declines towards the longest bin, while the excess at no recourse remains positive in bins where models make errors. In contrast, maximally split predictions ($4$ or $5$ correct predictions) occur roughly as often as the baseline suggests below the median and fall further below it above the median. \autoref{fig:appendix-length-vote-excess-joint} reports the excess for every recourse level, and \autoref{fig:appendix-length-recourse-by-label} shows that the pattern holds also when conditioning on the true label.

\paragraph{Conclusions.}

Individuals eliciting short traces receive near-unanimous predictions, whereas those eliciting longer traces receive less consistent predictions, even after accounting for accuracy. Yet among individuals eliciting the longest traces, all nine models misclassify $5$--$10\%$, exceeding the independent-error baseline. Greater dependence on model choice therefore coexists with shared failure.

\section{Discussion}\label{sec:discussion}
Does more reasoning push predictions towards monoculture or multiplicity? Our results show how these tendencies can coexist. Enabling reasoning increases agreement beyond an accuracy-based baseline when the same models evaluate the same individuals, while longer observed traces identify groups receiving less consistent predictions within fixed reasoning settings.
The association between longer traces and lower agreement extends prior observations of within-model inconsistency \citep{hagele2026thehotmess} to agreement between models. The intervention and trace-length analyses answer different questions: one concerns how enabling reasoning changes agreement, the other how disagreement varies with observed reasoning length. Observing longer traces therefore does not imply that increasing reasoning would make predictions more diverse.

The individual-level results show why agreement alone is insufficient to assess access to correct predictions. Enabling reasoning concentrates outcomes at both unanimous correctness and unanimous error. Among individuals whose recourse changes, increases outnumber decreases. Nevertheless, some lose their sole source of a correct prediction, and the share misclassified by all nine models rises by 2.0-2.4 percentage points. Conversely, individuals eliciting longer traces receive more mixed predictions, but some remain misclassified by every model. Multiplicity across a group can thus coexist with systemic failure for individuals within it.

Our findings cover nine models from two developers and three binary prediction tasks. The intervention concerns reasoning at inference: disabling it does not undo the effects of reasoning training on output diversity \citep{karouzos2026WhereDoesOutput}. The length analysis compares groups of individuals and does not isolate the effect of extending a trace or establish length as a reliable predictor of disagreement for new individuals. Our estimates of excess agreement are also relative to an accuracy-based baseline. Accounting for item difficulty may change those estimates \citep{jo2026SubjectivityMonoculture}, while leaving the observed frequency of unanimous errors unchanged.  
Whether unanimous errors reflect model similarities or item difficulty, affected individuals lack a correct alternative within the evaluated model set.

For decisions about individuals, our findings motivate evaluating access to correct alternatives alongside average accuracy and agreement.\ Greater agreement does not necessarily mean greater reliability, and access to multiple models does not guarantee access to a correct prediction.\ Our recourse measure captures the availability of correct predictions within the evaluated model set; it does not establish whether an individual can identify or access a model providing one.
Although our experiments do not measure deployment harms, they show how shared errors can limit recourse within an evaluated model set. Future work should extend the analysis to more diverse models and decision settings, and compare increased reasoning effort , repeated sampling, and model switching, with and without answer aggregation. Examining which individuals benefit from each approach could clarify when switching models offers meaningful alternatives that additional computation within one model does not.

\subsubsection*{Acknowledgments}
We thank Yatong Chen, Vivian Nastl, and Mina Remeli for valuable feedback on earlier versions of this paper. We particularly thank Ricardo Dominguez-Olmedo for his support during the final stages of this work. The authors thank the International Max Planck Research School for Intelligent Systems (IMPRS-IS) for supporting Mila Gorecki.

\bibliography{literature}

@misc{anthropic2026effortdocs,
  title = {Effort},
  author = {{Anthropic}},
  year = 2026,
  journal = {Claude Platform Docs},
  urldate = {2026-09-21},
  howpublished = {Claude Platform Documentation},
  langid = {english}
}

@article{ballon2026RelationshipReasoningPerformance,
  title = {The Relationship between Reasoning and Performance in Large Language Models---O3 (Mini) Thinks Harder, Not Longer},
  author = {Ballon, Marthe and Algaba, Andres and Ginis, Vincent},
  year = 2026,
  month = may,
  journal = {Scientific Reports},
  volume = {16},
  number = {1},
  pages = {20896},
  publisher = {Nature Publishing Group},
  issn = {2045-2322},
  doi = {10.1038/s41598-026-50923-2},
  urldate = {2026-09-12},
  copyright = {2026 The Author(s)},
  langid = {english}
}

@misc{becker1996AdultDataset,
  title = {Adult {{Dataset}}},
  author = {Becker, Barry and Kohavi, Ronny},
  year = 1996,
  publisher = {UCI Machine Learning Repository},
  doi = {10.24432/C5XW20}
}

@inproceedings{black2022ModelMultiplicityOpportunities,
  title = {Model {{Multiplicity}}: {{Opportunities}}, {{Concerns}}, and {{Solutions}}},
  shorttitle = {Model {{Multiplicity}}},
  booktitle = {Proceedings of the 2022 {{ACM Conference}} on {{Fairness}}, {{Accountability}}, and {{Transparency}}},
  author = {Black, Emily and Raghavan, Manish and Barocas, Solon},
  year = 2022,
  month = jun,
  series = {{{FAccT}} '22},
  pages = {850--863},
  publisher = {Association for Computing Machinery},
  address = {New York, NY, USA},
  doi = {10.1145/3531146.3533149},
  urldate = {2024-10-02},
  isbn = {978-1-4503-9352-2}
}

@misc{bommasani2022OpportunitiesRisksFoundation,
  title = {On the {{Opportunities}} and {{Risks}} of {{Foundation Models}}},
  author = {Bommasani, Rishi and Hudson, Drew A. and Adeli, Ehsan and Altman, Russ and Arora, Simran and {von Arx}, Sydney and Bernstein, Michael S. and Bohg, Jeannette and Bosselut, Antoine and Brunskill, Emma and Brynjolfsson, Erik and Buch, Shyamal and Card, Dallas and Castellon, Rodrigo and Chatterji, Niladri and others},
  year = 2022,
  month = jul,
  number = {arXiv:2108.07258},
  eprint = {2108.07258},
  primaryclass = {cs},
  publisher = {arXiv},
  urldate = {2023-03-30},
  archiveprefix = {arXiv}
}

@inproceedings{bommasani2022PickingSamePerson,
  title = {Picking on the {{Same Person}}: {{Does Algorithmic Monoculture}} Lead to {{Outcome Homogenization}}?},
  shorttitle = {Picking on the {{Same Person}}},
  booktitle = {Advances in {{Neural Information Processing Systems}} 35},
  author = {Bommasani, Rishi and Creel, Kathleen A. and Kumar, Ananya and Jurafsky, Dan and Liang, Percy},
  year = 2022,
  month = nov,
  volume = {35},
  eprint = {2211.13972},
  primaryclass = {cs},
  pages = {3663--3678},
  publisher = {arXiv},
  doi = {10.48550/arXiv.2211.13972},
  urldate = {2024-10-02},
  archiveprefix = {arXiv},
  isbn = {978-1-7138-7108-8}
}

@inproceedings{bommasani2026AlgorithmicMonoculturesHiring,
  title = {Algorithmic {{Monocultures}} in {{Hiring}}},
  booktitle = {Proceedings of the 2026 {{ACM Conference}} on {{Fairness}}, {{Accountability}}, and {{Transparency}}},
  author = {Bommasani, Rishi and Bana, Sarah H. and Creel, Kathleen A. and Jurafsky, Dan and Liang, Percy},
  year = 2026,
  month = jun,
  series = {{{FAccT}} '26},
  pages = {6351--6382},
  publisher = {Association for Computing Machinery},
  address = {New York, NY, USA},
  doi = {10.1145/3805689.3812400},
  isbn = {979-8-4007-2596-8},
  langid = {english}
}

@article{breiman2001StatisticalModelingTwo,
  title = {Statistical {{Modeling}}: {{The Two Cultures}}},
  shorttitle = {Statistical {{Modeling}}},
  author = {Breiman, Leo},
  year = 2001,
  journal = {Statistical Science},
  volume = {16},
  number = {3},
  eprint = {2676681},
  eprinttype = {jstor},
  pages = {199--215},
  publisher = {Institute of Mathematical Statistics},
  issn = {0883-4237},
  urldate = {2024-12-02}
}

@article{byrt1993BiasPrevalenceKappa,
  title = {Bias, Prevalence and Kappa},
  author = {Byrt, Ted and Bishop, Janet and Carlin, John B.},
  year = 1993,
  journal = {Journal of Clinical Epidemiology},
  volume = {46},
  number = {5},
  pages = {423--429},
  issn = {0895-4356},
  doi = {10.1016/0895-4356(93)90018-V},
  copyright = {https://www.elsevier.com/tdm/userlicense/1.0/},
  langid = {english}
}

@inproceedings{chen2022HAPILargescaleLongitudinal,
  title = {{{HAPI}}: A Large-Scale Longitudinal Dataset of Commercial {{ML API}} Predictions},
  shorttitle = {{{HAPI}}},
  booktitle = {Proceedings of the 36th {{International Conference}} on {{Neural Information Processing Systems}}},
  author = {Chen, Lingjiao and Jin, Zhihua and Eyuboglu, Sabri and R{\'e}, Christopher and Zaharia, Matei and Zou, James},
  year = 2022,
  month = nov,
  series = {{{NIPS}} '22},
  pages = {24571--24585},
  publisher = {Curran Associates Inc.},
  address = {Red Hook, NY, USA},
  urldate = {2025-01-29},
  isbn = {978-1-7138-7108-8}
}

@misc{comanici2025Gemini25Pushing,
  title = {Gemini 2.5: {{Pushing}} the {{Frontier}} with {{Advanced Reasoning}}, {{Multimodality}}, {{Long Context}}, and {{Next Generation Agentic Capabilities}}},
  shorttitle = {Gemini 2.5},
  author = {Comanici, Gheorghe and Bieber, Eric and Schaekermann, Mike and Pasupat, Ice and Sachdeva, Noveen and Dhillon, Inderjit and Blistein, Marcel and Ram, Ori and Zhang, Dan and Rosen, Evan and Marris, Luke and Petulla, Sam and Gaffney, Colin and Aharoni, Asaf and Lintz, Nathan and others},
  year = 2025,
  month = dec,
  number = {arXiv:2507.06261},
  eprint = {2507.06261},
  primaryclass = {cs.CL},
  publisher = {arXiv},
  doi = {10.48550/arXiv.2507.06261},
  urldate = {2026-09-25},
  archiveprefix = {arXiv}
}

@inproceedings{creel2021AlgorithmicLeviathanArbitrariness,
  title = {The {{Algorithmic Leviathan}}: {{Arbitrariness}}, {{Fairness}}, and {{Opportunity}} in {{Algorithmic Decision Making Systems}}},
  shorttitle = {The {{Algorithmic Leviathan}}},
  booktitle = {Proceedings of the 2021 {{ACM Conference}} on {{Fairness}}, {{Accountability}}, and {{Transparency}}},
  author = {Creel, Kathleen and Hellman, Deborah},
  year = 2021,
  month = mar,
  series = {{{FAccT}} '21},
  pages = {816},
  publisher = {Association for Computing Machinery},
  address = {New York, NY, USA},
  doi = {10.1145/3442188.3445942},
  urldate = {2024-12-03},
  isbn = {978-1-4503-8309-7}
}

@inproceedings{cruz2024EvaluatingLanguageModels,
  title = {Evaluating Language Models as Risk Scores},
  booktitle = {Advances in {{Neural Information Processing Systems}} ({{Datasets}} and {{Benchmarks Track}})},
  author = {Cruz, Andr{\'e} F. and Hardt, Moritz and {Mendler-D{\"u}nner}, Celestine},
  year = 2024,
  volume = {37},
  eprint = {2407.14614},
  archiveprefix = {arXiv},
  langid = {english}
}

@article{damour2020UnderspecificationPresentsChallenges,
  title = {Underspecification {{Presents Challenges}} for {{Credibility}} in {{Modern Machine Learning}}},
  author = {D'Amour, Alexander and Heller, Katherine and Moldovan, Dan and Adlam, Ben and Alipanahi, Babak and Beutel, Alex and Chen, Christina and Deaton, Jonathan and Eisenstein, Jacob and Hoffman, Matthew D. and Hormozdiari, Farhad and Houlsby, Neil and Hou, Shaobo and Jerfel, Ghassen and Karthikesalingam, Alan and others},
  year = 2020,
  month = nov,
  journal = {Journal of Machine Learning Research},
  volume = {23},
  number = {226},
  eprint = {2011.03395},
  primaryclass = {cs, stat},
  pages = {1--61},
  publisher = {arXiv},
  issn = {1532-4435},
  doi = {10.48550/arXiv.2011.03395},
  urldate = {2024-10-02},
  archiveprefix = {arXiv},
  langid = {english}
}

@article{deepseek-ai2025DeepSeekR1IncentivizingReasoning,
  title = {{{DeepSeek-R1}}: {{Incentivizing Reasoning Capability}} in {{LLMs}} via {{Reinforcement Learning}}},
  shorttitle = {{{DeepSeek-R1}}},
  author = {{DeepSeek-AI} and Guo, Daya and Yang, Dejian and Zhang, Haowei and Song, Junxiao and Wang, Peiyi and Zhu, Qihao and Xu, Runxin and Zhang, Ruoyu and Ma, Shirong and Bi, Xiao and Zhang, Xiaokang and Yu, Xingkai and Wu, Yu and Wu, Z. F. and others},
  year = 2025,
  month = sep,
  journal = {Nature},
  volume = {645},
  number = {8081},
  eprint = {2501.12948},
  primaryclass = {cs},
  pages = {633--638},
  issn = {0028-0836, 1476-4687},
  doi = {10.1038/s41586-025-09422-z},
  urldate = {2026-01-07},
  archiveprefix = {arXiv}
}

@misc{devic2025TraceLengthSimple,
  title = {Trace {{Length}} Is a {{Simple Uncertainty Signal}} in {{Reasoning Models}}},
  author = {Devic, Siddartha and Peale, Charlotte and Bradley, Arwen and Williamson, Sinead and Nakkiran, Preetum and Gollakota, Aravind},
  year = 2025,
  month = oct,
  number = {arXiv:2510.10409},
  eprint = {2510.10409},
  primaryclass = {cs},
  publisher = {arXiv},
  doi = {10.48550/arXiv.2510.10409},
  urldate = {2026-05-04},
  archiveprefix = {arXiv}
}

@inproceedings{ding2022RetiringAdultNew,
  title = {Retiring {{Adult}}: {{New Datasets}} for {{Fair Machine Learning}}},
  shorttitle = {Retiring {{Adult}}},
  booktitle = {Advances in {{Neural Information Processing Systems}}},
  author = {Ding, Frances and Hardt, Moritz and Miller, John and Schmidt, Ludwig},
  year = 2022,
  month = jan,
  volume = {34},
  eprint = {2108.04884},
  primaryclass = {cs, stat},
  pages = {6478--6490},
  urldate = {2022-10-18},
  archiveprefix = {arXiv},
  langid = {english}
}

@article{feinstein1990HighAgreementLow,
  title = {High Agreement but Low Kappa: {{I}}. {{The}} Problems of Two Paradoxes},
  shorttitle = {High Agreement but Low {{Kappa}}},
  author = {Feinstein, Alvan R. and Cicchetti, Domenic V.},
  year = 1990,
  journal = {Journal of Clinical Epidemiology},
  volume = {43},
  number = {6},
  pages = {543--549},
  issn = {0895-4356},
  doi = {10.1016/0895-4356(90)90158-L},
  copyright = {https://www.elsevier.com/tdm/userlicense/1.0/},
  langid = {english}
}

@article{fisher2019AllModelsAre,
  title = {All {{Models}} Are {{Wrong}}, but {{Many}} Are {{Useful}}: {{Learning}} a {{Variable}}'s {{Importance}} by {{Studying}} an {{Entire Class}} of {{Prediction Models Simultaneously}}},
  shorttitle = {All {{Models}} Are {{Wrong}}, but {{Many}} Are {{Useful}}},
  author = {Fisher, Aaron and Rudin, Cynthia and Dominici, Francesca},
  year = 2019,
  journal = {Journal of Machine Learning Research},
  volume = {20},
  number = {177},
  pages = {1--81},
  issn = {1533-7928},
  urldate = {2024-11-14}
}

@inproceedings{ganesh2025SystemizingMultiplicityCurious,
  title = {Systemizing {{Multiplicity}}: {{The Curious Case}} of {{Arbitrariness}} in {{Machine Learning}}},
  booktitle = {Proceedings of the {{AAAI}}/{{ACM Conference}} on {{AI}}, {{Ethics}}, and {{Society}}},
  author = {Ganesh, Prakhar and Taik, Afaf and Farnadi, Golnoosh},
  year = 2025,
  volume = {8},
  eprint = {2501.14959},
  pages = {1032--1048},
  issn = {3065-8365},
  doi = {10.1609/aies.v8i2.36609},
  archiveprefix = {arXiv}
}

@inproceedings{geirhos2020AccuracyQuantifyingTrialbytrial,
  title = {Beyond Accuracy: {{Quantifying}} Trial-by-Trial Behaviour of {{CNNs}} and Humans by Measuring Error Consistency},
  shorttitle = {Beyond Accuracy},
  booktitle = {Proceedings of the 34th {{International Conference}} on {{Neural Information Processing Systems}}},
  author = {Geirhos, Robert and Meding, Kristof and Wichmann, Felix A.},
  year = 2020,
  volume = {33},
  eprint = {2006.16736},
  pages = {13890--13902},
  archiveprefix = {arXiv},
  isbn = {978-1-7138-2954-6}
}

@article{gema2025InverseScalingTestTime,
  title = {Inverse {{Scaling}} in {{Test-Time Compute}}},
  author = {Gema, Aryo Pradipta and H{\"a}gele, Alexander and Chen, Runjin and Arditi, Andy and {Goldman-Wetzler}, Jacob and {Fraser-Taliente}, Kit and Sleight, Henry and Petrini, Linda and Michael, Julian and Alex, Beatrice and Minervini, Pasquale and Chen, Yanda and Benton, Joe and Perez, Ethan},
  year = 2025,
  month = dec,
  journal = {Transactions on Machine Learning Research},
  issn = {2835-8856},
  langid = {english}
}

@inproceedings{goel2025GreatModelsThink,
  title = {Great {{Models Think Alike}} and This {{Undermines AI Oversight}}},
  booktitle = {Proceedings of the 42nd International Conference on Machine Learning},
  author = {Goel, Shashwat and Struber, Joschka and Auzina, Ilze Amanda and Chandra, Karuna K. and Kumaraguru, Ponnurangam and Kiela, Douwe and Prabhu, Ameya and Bethge, Matthias and Geiping, Jonas},
  year = 2025,
  month = jun,
  series = {Proceedings of Machine Learning Research},
  volume = {267},
  eprint = {2502.04313},
  primaryclass = {cs},
  pages = {19621--19678},
  publisher = {PMLR},
  doi = {10.48550/arXiv.2502.04313},
  urldate = {2026-03-23},
  archiveprefix = {arXiv},
  langid = {english}
}

@inproceedings{gorecki2025MonocultureMultiplicityWhich,
  title = {Monoculture or {{Multiplicity}}: {{Which Is It}}?},
  shorttitle = {Monoculture or {{Multiplicity}}},
  booktitle = {The {{Thirty}}-Ninth {{Annual Conference}} on {{Neural Information Processing Systems}}},
  author = {Gorecki, Mila and Hardt, Moritz},
  year = 2025,
  month = oct,
  urldate = {2026-01-08},
  langid = {english}
}

@inproceedings{gur-arieh2025ConsistentlyArbitraryArbitrarily,
  title = {Consistently {{Arbitrary}} or {{Arbitrarily Consistent}}: {{Navigating}} the {{Tensions Between Homogenization}} and {{Multiplicity}} in {{Algorithmic Decision-Making}}},
  shorttitle = {Consistently {{Arbitrary}} or {{Arbitrarily Consistent}}},
  booktitle = {Proceedings of the 2025 {{ACM Conference}} on {{Fairness}}, {{Accountability}}, and {{Transparency}}},
  author = {{Gur-Arieh}, Shira and Lee, Christina},
  year = 2025,
  month = jun,
  series = {{{FAccT}} '25},
  pages = {3336--3349},
  publisher = {Association for Computing Machinery},
  address = {New York, NY, USA},
  doi = {10.1145/3715275.3732215},
  urldate = {2025-06-25},
  isbn = {979-8-4007-1482-5}
}

@inproceedings{gur-arieh2026AmbiguityCollapseLLMs,
  title = {Ambiguity {{Collapse}} by {{LLMs}}: {{A Taxonomy}} of {{Epistemic Risks}}},
  booktitle = {Proceedings of the 2026 {{ACM Conference}} on {{Fairness}}, {{Accountability}}, and {{Transparency}}},
  author = {{Gur-Arieh}, Shira and Wang, Angelina and Fazelpour, Sina},
  year = 2026,
  series = {{{FAccT}} '26},
  pages = {5959--5980},
  publisher = {Association for Computing Machinery},
  address = {New York, NY, USA},
  doi = {10.1145/3805689.3806469},
  isbn = {979-8-4007-2596-8}
}

@inproceedings{hagele2026thehotmess,
  title = {The Hot Mess of {{AI}}: {{How}} Does Misalignment Scale with Model Intelligence and Task Complexity?},
  booktitle = {The Fourteenth International Conference on Learning Representations},
  author = {H{\"a}gele, Alexander and Gema, Aryo Pradipta and Sleight, Henry and Perez, Ethan and {Sohl-Dickstein}, Jascha},
  year = 2026
}

@misc{han2025TokenBudgetAwareLLMReasoning,
  title = {Token-{{Budget-Aware LLM Reasoning}}},
  author = {Han, Tingxu and Wang, Zhenting and Fang, Chunrong and Zhao, Shiyu and Ma, Shiqing and Chen, Zhenyu},
  year = 2025,
  month = jun,
  number = {arXiv:2412.18547},
  eprint = {2412.18547},
  primaryclass = {cs.CL},
  publisher = {arXiv},
  doi = {10.48550/arXiv.2412.18547},
  urldate = {2026-09-25},
  archiveprefix = {arXiv}
}

@inproceedings{hardt2022BackwardBaselinesYour,
  title = {Is {{Your Model Predicting}} the {{Past}}?},
  booktitle = {Proceedings of the 3rd {{ACM}} Conference on Equity and Access in Algorithms, Mechanisms, and Optimization},
  author = {Kim, Michael P. and Hardt, Moritz},
  year = 2023,
  series = {{{EAAMO}} '23},
  pages = {1--8},
  publisher = {Association for Computing Machinery},
  doi = {10.1145/3617694.3623225},
  isbn = {979-8-4007-0381-2},
  langid = {english}
}

@misc{hedden2026AlgorithmicMonocultureIts,
  title = {Algorithmic {{Monoculture}} and Its {{Critics}}},
  author = {Hedden, Brian and Raghavan, Manish},
  year = 2026,
  month = apr,
  number = {arXiv:2604.06047},
  eprint = {2604.06047},
  primaryclass = {cs.CY},
  publisher = {arXiv},
  doi = {10.48550/arXiv.2604.06047},
  urldate = {2026-06-18},
  archiveprefix = {arXiv}
}

@inproceedings{jain2024AlgorithmicPluralismStructural,
  title = {Algorithmic {{Pluralism}}: {{A Structural Approach To Equal Opportunity}}},
  shorttitle = {Algorithmic {{Pluralism}}},
  booktitle = {The 2024 {{ACM Conference}} on {{Fairness}}, {{Accountability}}, and {{Transparency}}},
  author = {Jain, Shomik and Suriyakumar, Vinith and Creel, Kathleen and Wilson, Ashia},
  year = 2024,
  month = jun,
  eprint = {2305.08157},
  primaryclass = {cs},
  pages = {197--206},
  doi = {10.1145/3630106.3658899},
  urldate = {2024-10-02},
  archiveprefix = {arXiv}
}

@misc{jain2026TaskDependentEvaluationLLM,
  title = {Task-{{Dependent Evaluation}} of {{LLM Output Homogenization}}: {{A Taxonomy-Guided Framework}}},
  shorttitle = {Task-{{Dependent Evaluation}} of {{LLM Output Homogenization}}},
  author = {Jain, Shomik and Lanchantin, Jack and Nickel, Maximilian and Ross, Candace and Ullrich, Karen and Wilson, Ashia and {Watson-Daniels}, Jamelle},
  year = 2026,
  month = apr,
  number = {arXiv:2509.21267},
  eprint = {2509.21267},
  primaryclass = {cs.CL},
  publisher = {arXiv},
  doi = {10.48550/arXiv.2509.21267},
  urldate = {2026-09-19},
  archiveprefix = {arXiv}
}

@inproceedings{jiang2025ArtificialHivemindOpenEnded,
  title = {Artificial {{Hivemind}}: {{The Open-Ended Homogeneity}} of {{Language Models}} (and {{Beyond}})},
  shorttitle = {Artificial {{Hivemind}}},
  booktitle = {The Thirty-Ninth Annual Conference on Neural Information Processing Systems Datasets and Benchmarks Track},
  author = {Jiang, Liwei and Chai, Yuanjun and Li, Margaret and Liu, Mickel and Fok, Raymond and Dziri, Nouha and Tsvetkov, Yulia and Sap, Maarten and Albalak, Alon and Choi, Yejin},
  year = 2025,
  langid = {english}
}

@misc{jo2026SubjectivityMonoculture,
  title = {The {{Subjectivity}} of {{Monoculture}}},
  author = {Jo, Nathanael and Garg, Nikhil and Raghavan, Manish},
  year = 2026,
  publisher = {arXiv},
  doi = {10.48550/ARXIV.2602.24086},
  urldate = {2026-03-02},
  copyright = {Creative Commons Attribution 4.0 International},
  langid = {english}
}

@misc{karouzos2026WhereDoesOutput,
  title = {Where Does Output Diversity Collapse in Post-Training?},
  author = {Karouzos, Constantinos and Tan, Xingwei and Aletras, Nikolaos},
  year = 2026,
  month = apr,
  number = {arXiv:2604.16027},
  eprint = {2604.16027},
  primaryclass = {cs.CL},
  publisher = {arXiv},
  doi = {10.48550/arXiv.2604.16027},
  urldate = {2026-09-19},
  archiveprefix = {arXiv}
}

@inproceedings{kim2025CorrelatedErrorsLarge,
  title = {Correlated {{Errors}} in {{Large Language Models}}},
  booktitle = {Proceedings of the 42nd {{International Conference}} on {{Machine Learning}}},
  author = {Kim, Elliot and Garg, Avi and Peng, Kenny and Garg, Nikhil},
  year = 2025,
  month = jun,
  series = {Proceedings of {{Machine Learning Research}}},
  volume = {267},
  eprint = {2506.07962},
  primaryclass = {cs},
  pages = {30038--30066},
  publisher = {arXiv},
  doi = {10.48550/arXiv.2506.07962},
  urldate = {2025-09-05},
  archiveprefix = {arXiv}
}

@article{kleinberg2021AlgorithmicMonocultureSocial,
  title = {Algorithmic Monoculture and Social Welfare},
  author = {Kleinberg, Jon and Raghavan, Manish},
  year = 2021,
  month = jun,
  journal = {Proceedings of the National Academy of Sciences},
  volume = {118},
  number = {22},
  pages = {e2018340118},
  issn = {0027-8424, 1091-6490},
  doi = {10.1073/pnas.2018340118},
  urldate = {2023-02-14},
  langid = {english}
}

@misc{kleinberg2026PriceAnarchyAlgorithmic,
  title = {Price of {{Anarchy}} of {{Algorithmic Monoculture}}},
  author = {Kleinberg, Robert and Sinanaj, Erald and Tardos, {\'E}va},
  year = 2026,
  month = apr,
  number = {arXiv:2604.00444},
  eprint = {2604.00444},
  primaryclass = {cs},
  publisher = {arXiv},
  doi = {10.48550/arXiv.2604.00444},
  urldate = {2026-04-02},
  archiveprefix = {arXiv}
}

@inproceedings{kojima2022LargeLanguageModels,
  title = {Large Language Models Are Zero-Shot Reasoners},
  booktitle = {Proceedings of the 36th {{International Conference}} on {{Neural Information Processing Systems}}},
  author = {Kojima, Takeshi and Gu, Shixiang Shane and Reid, Machel and Matsuo, Yutaka and Iwasawa, Yusuke},
  year = 2022,
  month = nov,
  series = {{{NIPS}} '22},
  pages = {22199--22213},
  publisher = {Curran Associates Inc.},
  address = {Red Hook, NY, USA},
  urldate = {2026-09-25},
  isbn = {978-1-7138-7108-8}
}

@misc{lee2025HowWellLLMs,
  title = {How {{Well}} Do {{LLMs Compress Their Own Chain}}-of-{{Thought}}? {{A Token Complexity Approach}}},
  shorttitle = {How {{Well}} Do {{LLMs Compress Their Own Chain}}-of-{{Thought}}?},
  author = {Lee, Ayeong and Che, Ethan and Peng, Tianyi},
  year = 2025,
  month = apr,
  number = {arXiv:2503.01141},
  eprint = {2503.01141},
  primaryclass = {cs.CL},
  publisher = {arXiv},
  doi = {10.48550/arXiv.2503.01141},
  urldate = {2026-09-25},
  archiveprefix = {arXiv}
}

@misc{liu2026LLMsJuryCrossModel,
  title = {{{LLMs}} as a {{Jury}}: {{Cross-Model Consensus Can Outperform Process Reward Models}} for {{LLM Reasoning}}},
  shorttitle = {{{LLMs}} as a {{Jury}}},
  author = {Liu, Ning},
  year = 2026,
  month = aug,
  publisher = {arXiv},
  doi = {10.48550/ARXIV.2607.10139},
  urldate = {2026-09-15},
  copyright = {Creative Commons Attribution 4.0 International}
}

@inproceedings{mania2019ModelSimilarityMitigates,
  title = {Model {{Similarity Mitigates Test Set Overuse}}},
  booktitle = {Advances in {{Neural Information Processing Systems}}},
  author = {Mania, Horia and Miller, John and Schmidt, Ludwig and Hardt, Moritz and Recht, Benjamin},
  year = 2019,
  volume = {32},
  publisher = {Curran Associates, Inc.},
  urldate = {2025-02-11}
}

@inproceedings{marx2020PredictiveMultiplicityClassification,
  title = {Predictive {{Multiplicity}} in {{Classification}}},
  booktitle = {Proceedings of the 37th {{International Conference}} on {{Machine Learning}}},
  author = {Marx, Charles and Calmon, Flavio and Ustun, Berk},
  year = 2020,
  month = nov,
  pages = {6765--6774},
  publisher = {PMLR},
  issn = {2640-3498},
  urldate = {2024-10-02},
  langid = {english}
}

@inproceedings{mei2026ReasoningUncertaintyReasoning,
  title = {Reasoning about {{Uncertainty}}: {{Do Reasoning Models Know When They Don}}'t {{Know}}?},
  shorttitle = {Reasoning about {{Uncertainty}}},
  booktitle = {Findings of the {{Association}} for {{Computational Linguistics}}: {{EACL}} 2026},
  author = {Mei, Zhiting and Zhang, Christina and Yin, Tenny and Lidard, Justin and Sho, Ola and Majumdar, Anirudha},
  editor = {Demberg, Vera and Inui, Kentaro and Marquez, Llu{\'i}s},
  year = 2026,
  month = mar,
  pages = {3408--3458},
  publisher = {Association for Computational Linguistics},
  address = {Rabat, Morocco},
  doi = {10.18653/v1/2026.findings-eacl.178},
  urldate = {2026-04-17},
  isbn = {979-8-89176-386-9}
}

@inproceedings{muennighoff2025S1SimpleTesttime,
  title = {S1: {{Simple}} Test-Time Scaling},
  shorttitle = {S1},
  booktitle = {Proceedings of the 2025 {{Conference}} on {{Empirical Methods}} in {{Natural Language Processing}}},
  author = {Muennighoff, Niklas and Yang, Zitong and Shi, Weijia and Li, Xiang Lisa and {Fei-Fei}, Li and Hajishirzi, Hannaneh and Zettlemoyer, Luke and Liang, Percy and Cand{\`e}s, Emmanuel and Hashimoto, Tatsunori},
  editor = {Christodoulopoulos, Christos and Chakraborty, Tanmoy and Rose, Carolyn and Peng, Violet},
  year = 2025,
  month = nov,
  pages = {20275--20321},
  publisher = {Association for Computational Linguistics},
  address = {Suzhou, China},
  doi = {10.18653/v1/2025.emnlp-main.1025},
  urldate = {2026-09-12},
  isbn = {979-8-89176-332-6}
}

@article{nayab2026ConciseThoughtsImpact,
  title = {Concise {{Thoughts}}: {{Impact}} of {{Output Length}} on {{LLM Reasoning}} and {{Cost}}},
  shorttitle = {Concise {{Thoughts}}},
  author = {Nayab, Sania and Rossolini, Giulio and Simoni, Marco and Saracino, Andrea and Buttazzo, Giorgio and Manes, Nicolamaria and Giacomelli, Fabrizio},
  year = 2026,
  month = dec,
  journal = {Information Sciences},
  volume = {757},
  eprint = {2407.19825},
  primaryclass = {cs.CL},
  pages = {123939},
  issn = {00200255},
  doi = {10.1016/j.ins.2026.123939},
  urldate = {2026-09-25},
  archiveprefix = {arXiv}
}

@misc{openai2024OpenAIO1System,
  title = {{{OpenAI}} O1 {{System Card}}},
  author = {OpenAI and Jaech, Aaron and Kalai, Adam and Lerer, Adam and Richardson, Adam and {El-Kishky}, Ahmed and Low, Aiden and Helyar, Alec and Madry, Aleksander and Beutel, Alex and Carney, Alex and Iftimie, Alex and Karpenko, Alex and Passos, Alex Tachard and Neitz, Alexander and others},
  year = 2024,
  month = dec,
  journal = {arXiv.org},
  urldate = {2026-09-15},
  howpublished = {https://arxiv.org/abs/2412.16720v2},
  langid = {english}
}

@techreport{openai2025GPT51SystemCard,
  title = {{{GPT-5}}.1 Instant and {{GPT-5}}.1 Thinking System Card Addendum},
  author = {{OpenAI}},
  year = 2025,
  month = nov,
  institution = {OpenAI}
}

@techreport{openai2025GPT52SystemCard,
  title = {Update to {{GPT-5}} System Card: {{GPT-5}}.2},
  author = {{OpenAI}},
  year = 2025,
  month = dec,
  institution = {OpenAI}
}

@misc{openai2025reasoningeffort,
  title = {Reasoning Models (Reasoning\_effort Parameter)},
  author = {{OpenAI}},
  year = 2025,
  urldate = {2026-09-21},
  howpublished = {OpenAI API Documentation},
  langid = {english}
}

@techreport{openai2026GPT54SystemCard,
  title = {{{GPT-5}}.4 Thinking System Card},
  author = {{OpenAI}},
  year = 2026,
  month = mar,
  institution = {OpenAI}
}

@misc{qwen2026Qwen35,
  title = {Qwen3.5: {{Towards}} Native Multimodal Agents},
  author = {{Qwen Team}},
  year = 2026,
  month = feb
}

@inproceedings{renze2024BenefitsConciseChaina,
  title = {The {{Benefits}} of a {{Concise Chain}} of {{Thought}} on {{Problem-Solving}} in {{Large Language Models}}},
  booktitle = {2024 2nd {{International Conference}} on {{Foundation}} and {{Large Language Models}} ({{FLLM}})},
  author = {Renze, Matthew and Guven, Erhan},
  year = 2024,
  month = nov,
  eprint = {2401.05618},
  primaryclass = {cs.CL},
  pages = {476--483},
  doi = {10.1109/FLLM63129.2024.10852493},
  urldate = {2026-09-25},
  archiveprefix = {arXiv}
}

@inproceedings{semenova2022ExistenceSimplerMachine,
  title = {On the {{Existence}} of {{Simpler Machine Learning Models}}},
  booktitle = {Proceedings of the 2022 {{ACM Conference}} on {{Fairness}}, {{Accountability}}, and {{Transparency}}},
  author = {Semenova, Lesia and Rudin, Cynthia and Parr, Ronald},
  year = 2022,
  month = jun,
  series = {{{FAccT}} '22},
  pages = {1827--1858},
  publisher = {Association for Computing Machinery},
  address = {New York, NY, USA},
  doi = {10.1145/3531146.3533232},
  urldate = {2025-07-04},
  isbn = {978-1-4503-9352-2}
}

@inproceedings{snell2024ScalingLLMTestTime,
  title = {Scaling {{LLM Test-Time Compute Optimally Can}} Be {{More Effective}} than {{Scaling Parameters}} for {{Reasoning}}},
  booktitle = {The {{Thirteenth International Conference}} on {{Learning Representations}}},
  author = {Snell, Charlie and Lee, Jaehoon and Xu, Kelvin and Kumar, Aviral},
  year = 2025,
  langid = {english}
}

@article{stechly2025SemanticsUnreasonableEffectiveness,
  title = {Beyond {{Semantics}}: {{The Unreasonable Effectiveness}} of {{Reasonless Intermediate Tokens}}},
  shorttitle = {Beyond {{Semantics}}},
  author = {Valmeekam, Karthik and Palod, Vardhan and Stechly, Kaya and Gundawar, Atharva and Kambhampati, Subbarao},
  year = 2026,
  journal = {Transactions on Machine Learning Research},
  issn = {2835-8856},
  langid = {english}
}

@inproceedings{toups2024EcosystemlevelAnalysisDeployed,
  title = {Ecosystem-Level Analysis of Deployed Machine Learning Reveals Homogeneous Outcomes},
  booktitle = {Proceedings of the 37th {{International Conference}} on {{Neural Information Processing Systems}}},
  author = {Toups, Connor and Bommasani, Rishi and Creel, Kathleen A. and Bana, Sarah H. and Jurafsky, Dan and Liang, Percy},
  year = 2024,
  month = may,
  series = {{{NIPS}} '23},
  pages = {51178--51201},
  publisher = {Curran Associates Inc.},
  address = {Red Hook, NY, USA},
  urldate = {2024-12-03}
}

@inproceedings{turpin2023LanguageModelsDont,
  title = {Language {{Models Don}}'t {{Always Say What They Think}}: {{Unfaithful Explanations}} in {{Chain}}-of-{{Thought Prompting}}},
  shorttitle = {Language {{Models Don}}'t {{Always Say What They Think}}},
  booktitle = {Advances in {{Neural Information Processing Systems}}},
  author = {Turpin, Miles and Michael, Julian and Perez, Ethan and Bowman, Samuel R.},
  year = 2023,
  month = dec,
  volume = {36},
  eprint = {2305.04388},
  primaryclass = {cs},
  publisher = {arXiv},
  doi = {10.48550/arXiv.2305.04388},
  urldate = {2026-05-04},
  archiveprefix = {arXiv},
  langid = {english}
}

@misc{u.s.censusbureau2014SurveyIncomeProgram,
  title = {Survey of {{Income}} and {{Program Participation}} ({{SIPP}})},
  author = {{U.S. Census Bureau}},
  year = 2014,
  urldate = {2025-07-28}
}

@misc{u.s.censusbureau2018AmericanCommunitySurvey,
  title = {American {{Community Survey}} ({{ACS}}) {{Public Use Microdata Sample}} ({{PUMS}})},
  author = {{U.S. Census Bureau}},
  year = 2018
}

@inproceedings{wang2023SELFCONSISTENCYIMPROVESCHAIN,
  title = {Self-{{Consistency Improves Chain}} of {{Thought Reasoning}} in {{Language Models}}},
  booktitle = {The {{Eleventh International Conference}} on {{Learning Representations}}},
  author = {Wang, Xuezhi and Wei, Jason and Schuurmans, Dale and Le, Quoc and Chi, Ed H and Narang, Sharan and Chowdhery, Aakanksha and Zhou, Denny},
  year = 2023,
  langid = {english}
}

@misc{wani2026ReasoningTaxToken,
  title = {The {{Reasoning Tax}}: {{Token Economics}} of {{LLM Reasoning Across Task Types}} and {{Deployment Contexts}}},
  shorttitle = {The {{Reasoning Tax}}},
  author = {Wani, Sachin Gopal and Dholakia, Ajay and Ellison, David},
  year = 2026,
  month = aug,
  number = {arXiv:2608.26235},
  eprint = {2608.26235},
  primaryclass = {cs.AI},
  publisher = {arXiv},
  doi = {10.48550/arXiv.2608.26235},
  urldate = {2026-09-25},
  archiveprefix = {arXiv}
}

@inproceedings{wei2022ChainofthoughtPromptingElicits,
  title = {Chain-of-Thought Prompting Elicits Reasoning in Large Language Models},
  booktitle = {Proceedings of the 36th {{International Conference}} on {{Neural Information Processing Systems}}},
  author = {Wei, Jason and Wang, Xuezhi and Schuurmans, Dale and Bosma, Maarten and Ichter, Brian and Xia, Fei and Chi, Ed H. and Le, Quoc V. and Zhou, Denny},
  year = 2022,
  month = nov,
  series = {{{NIPS}} '22},
  pages = {24824--24837},
  publisher = {Curran Associates Inc.},
  address = {Red Hook, NY, USA},
  urldate = {2026-09-15},
  isbn = {978-1-7138-7108-8}
}

@inproceedings{wu2024GenerativeMonocultureLarge,
  title = {Generative {{Monoculture}} in {{Large Language Models}}},
  booktitle = {The {{Thirteenth International Conference}} on {{Learning Representations}}},
  author = {Wu, Fan and Black, Emily and Chandrasekaran, Varun},
  year = 2025,
  langid = {english}
}

@techreport{yang2025Qwen3TechnicalReport,
  title = {Qwen3 Technical Report},
  author = {Yang, An and Li, Anfeng and Yang, Baosong and Zhang, Beichen and Hui, Binyuan and Zheng, Bo and Yu, Bowen and others},
  year = 2025,
  month = may,
  number = {arXiv:2505.09388},
  institution = {Qwen Team}
}

\appendix
\setcounter{figure}{0}\renewcommand{\thefigure}{A\arabic{figure}}\renewcommand{\theHfigure}{A\arabic{figure}}
\setcounter{table}{0}\renewcommand{\thetable}{A\arabic{table}}\renewcommand{\theHtable}{A\arabic{table}}
\section{Appendix}
\FloatBarrier
\subsection{Experimental Details}\label{appendix:resources}\label{appendix:compute}
\paragraph{Code and reproducibility.}
Code to reproduce the analysis is available at \url{https://github.com/socialfoundations/reasonance.git}. Appendix~\ref{appendix:tasks} describes the datasets and evaluation samples, Appendix~\ref{appendix:models} lists the evaluated models, and Appendix~\ref{appendix:inference-protocol} details the prompts, generation settings, resampling seeds, and response processing.

\paragraph{Compute resources and usage.}
pen-weight models were run with vLLM on a shared compute cluster, using one GPU per job; closed-weight models were accessed through provider APIs. Table~\ref{tab:appendix-compute} summarizes approximate inference resources for the numeric-QA runs of the six Qwen and three OpenAI models used in the reported analyses. Local inference used approximately $7{,}800$ GPU-hours. API inference used approximately $502$ million prompt tokens and $826$ million completion tokens.

\begin{table}[htbp]
\caption{Approximate inference resources by model group. GPU-hours cover local inference; API token counts are reported in millions (M). Dashes indicate quantities not reported for that inference backend.}
\label{tab:appendix-compute}
\centering
\begin{tabular}{lrrr}
\toprule
Model group & GPU-hours & Prompt tokens (M) & Completion tokens (M) \\
\midrule
Qwen (six models) & 7,800 & -- & -- \\
OpenAI (three models) & -- & 502 & 826 \\
\bottomrule
\end{tabular}
\end{table}

\FloatBarrier
\subsection{Tasks}\label{appendix:tasks}
We evaluate three binary classification tasks from two U.S. Census Bureau data sources: the American Community Survey (ACS) Public Use Microdata Sample (PUMS)~\citep{u.s.censusbureau2018AmericanCommunitySurvey} and the Survey of Income and Program Participation (SIPP)~\citep{u.s.censusbureau2014SurveyIncomeProgram}.

\paragraph{Evaluation samples.}
We randomly sample $8323$ instances from ACSIncome ($5\%$ of the full dataset) and $5000$ from ACSEmployment ($1.5\%$), and use all $3972$ SIPP test records ($10\%$ of the full dataset). Evaluation samples exclude the $2000$ validation individuals used for threshold fitting. All models and reasoning configurations receive the same inputs.

\paragraph{ACS prediction tasks.}
We use person-level data from the 2018 ACS one-year survey, adopting the ACSIncome and ACSEmployment task definitions, feature sets, and population filters from \texttt{folktables}~\citep{ding2022RetiringAdultNew}. We encode features using the natural-text mappings provided by \texttt{folktexts}~\citep{cruz2024EvaluatingLanguageModels}. Table~\ref{tab:folktexts_col_to_text} lists the features and example encodings. Prior work found a larger performance gap between language models and traditional machine learning models on these tasks~\citep{gorecki2025MonocultureMultiplicityWhich}.

\paragraph{ACSIncome.}
ACSIncome predicts whether annual income (PINCP) exceeds \$50,000 among employed U.S. residents older than 16. Its features are AGEP, COW, SCHL, MAR, OCCP, POBP, RELP, WKHP, SEX, and RAC1P. The task was proposed as a successor to the UCI Adult dataset~\citep{becker1996AdultDataset}, widely used in algorithmic fairness research.

\paragraph{ACSEmployment.}
ACSEmployment predicts employment status from the ESR target column for U.S. residents. Its features are AGEP, SCHL, MAR, RELP, DIS, ESP, CIT, MIG, MIL, ANC, NATIVITY, DEAR, DEYE, DREM, SEX, and RAC1P.

\paragraph{SIPP.}
SIPP is a longitudinal survey of income, employment, household composition, and program participation. Following \citet{hardt2022BackwardBaselinesYour}, we use Waves 1 and 2 of the 2014 panel and adopt their data cleaning and preprocessing. The task predicts whether an individual's income is well above the Official Poverty Measure (OPM), a cash-income-based measure of poverty, using the Wave 2 OPM ratio as the basis for the binary target. We construct 50 features from Wave 1 variables, covering demographics, income, health, hardship, and program participation, and convert them to natural text using the SIPP codebook. Table~\ref{tab:sipp_col_to_text} provides feature descriptions and example encodings.

\paragraph{SIPP preprocessing.}
We adapt the pipeline of \citet{hardt2022BackwardBaselinesYour}, retaining the data source, panel, wave split, and target definition. We correct personal income to avoid counting investment and property income twice, and exclude identifiers and the Wave 1 poverty measure from the predictors. We also preserve the codebook's living-quarter categories and revise feature descriptions for clarity. The prediction question explicitly refers to household income, consistent with the target's household income-to-poverty ratio. All reported results use this preprocessing.

\begin{landscape}
{   
    \begin{minipage}{0.95\linewidth}
    \footnotesize
    \captionsetup{hypcap=false}
    \captionof{table}{Description of column-to-text mappings for the ACS features used by ACSIncome or ACSEmployment. The variable part of each example is shown in  {\color{gray}\tt typeset gray font}. See the ACS PUMS data dictionary for the full list of available variables.\protect\footnotemark[1]} \label{tab:folktexts_col_to_text}
    \begin{tabular}{p{0.08\linewidth}p{0.25\linewidth}p{0.67\linewidth}}
        \toprule
        \textit{\textbf{Column}} & \textit{\textbf{Description}} & \textit{\textbf{Example}} \\
        \midrule
        AGEP & Age & age is {\color{gray}\tt 31 years old} \\
        COW & Class of worker &  class of worker is {\color{gray}\tt Working for a for-profit private company or organization} \\
        SCHL & Educational attainment & highest educational attainment is {\color{gray}\tt Bachelor's degree} \\
        MAR & Marital status & marital status is {\color{gray}\tt Married} \\
        OCCP & Occupation &  occupation is {\color{gray}\tt Cashiers} \\
        POBP & Place of birth & place of birth is {\color{gray}\tt California} \\
        RELP & Relationship & relationship to the reference survey respondent in the survey is {\color{gray}\tt The reference person itself} \\
        WKHP & Work-hours per week & usual number of hours worked per week is {\color{gray}\tt 32 hours} \\
        SEX & Sex & sex is {\color{gray}\tt Female} \\
        RAC1P & Race & race is {\color{gray}\tt Black or African American} \\
        DIS & Disability status & disability status is {\color{gray}\tt No disability}\\
        ESP & Employment status of parents & employment status of parents is {\color{gray}\tt N/A (not own child of householder, and not child in subfamily)}\\
        CIT & Citizenship status & citizenship status is {\color{gray}\tt Born in the United States} \\
        MIG & Mobility (lived here 1 year ago) & mobility status over the last year is {\color{gray}\tt Lived in a different house in the United States one year ago} \\
        MIL & Military service & military service status is {\color{gray}\tt Never served in the military} \\
        ANC & Ancestry & ancestry is {\color{gray}\tt Single ancestry} \\
        NATIVITY & Nativity & nativity is {\color{gray}\tt Native}\\
        DEAR & Hearing & hearing status is  {\color{gray}\tt No hearing difficulty} \\
        DEYE & Vision & vision status is {\color{gray}\tt With vision difficulty}. \\
        DREM & Cognition & cognition status is {\color{gray}\tt No cognitive difficulty} \\
        \bottomrule
    \end{tabular}
     \end{minipage}
     \footnotetext[1]{\url{https://www.census.gov/programs-surveys/acs/microdata/documentation.html}}
    }
\end{landscape}

\begin{landscape}
\footnotesize
\begin{center}
\begin{longtable}{p{0.3\linewidth}p{0.27\linewidth}p{0.43\linewidth}}
\caption{Description of all column-to-text mappings implemented for SIPP features. The variable part of each example is shown in  {\color{gray}\tt typeset gray font}. See the SIPP codebook and the pre-processing script provided by \citet{hardt2022BackwardBaselinesYour} for the full list of available variables.\protect\footnotemark[1]} \label{tab:sipp_col_to_text} \\
\toprule
\textit{\textbf{Column}} & \textit{\textbf{Description}} & \textit{\textbf{Example}} \\
\midrule
\endfirsthead

\multicolumn{3}{l}{\textit{(Continued from previous page)}} \\
\toprule
\textit{\textbf{Column}} & \textit{\textbf{Description}} & \textit{\textbf{Example}} \\
\midrule
\endhead

\midrule
\multicolumn{3}{l}{\textit{(Continued on next page)}} \\
\endfoot

\bottomrule
\endlastfoot

    \multicolumn{3}{l}{\textit{demographics}} \\
    AGE& age & age is {\color{gray}\tt 49 years old} \\
    GENDER& gender & gender is {\color{gray}\tt Male} \\
    RACE& race & self-identified race(s) is {\color{gray}\tt White only} \\
    EDUCATION& highest educational attainment & highest level of education completed is {\color{gray}\tt High School Graduate (diploma or GED or equivalent)} \\
    MARITAL\_STATUS& marital status & marital status is {\color{gray}\tt Married, spouse present} \\
    CITIZENSHIP\_STATUS& US citizenship status & US citizenship status is {\color{gray}\tt Yes} \\
    FAMILY\_SIZE\_AVG& average family size in the reference year & average number of persons in family is {\color{gray}\tt 4 persons} \\
    ORIGIN& of Spanish, Hispanic, or Latino origin & Spanish, Hispanic, or Latino origin is {\color{gray}\tt No} \\
    \multicolumn{3}{l}{\textit{income}} \\
    INCOME& total annual personal income from earnings, investment and property, transfers, social insurance, and other sources & total personal income from personal earnings, investment and property, means-based transfer income, social insurance payments and other sources in given year is {\color{gray}\tt \$14498} \\
    HOUSEHOLD\_INC& total household income in the reference year & total annual income of all household members is {\color{gray}\tt \$61422} \\
    RECEIVED\_WORK\_COMP& receives worker's compensation payments & receipt of worker's compensation payments is {\color{gray}\tt No} \\
    INCOME\_FROM\_ASSISTANCE& total annual income from transfers, social insurance, and other non-earned sources & total income from transfers, social insurance and other non-earned sources in given year is {\color{gray}\tt \$0} \\
    SAVINGS\_INV\_AMOUNT& total value of IRA, KEOGH, 401k, 403b, 503b, and Thrift Savings Plan accounts & total value of retirement accounts is {\color{gray}\tt \$0} \\
    \multicolumn{3}{l}{\textit{hardship}} \\
    LIVING\_QUARTERS\_TYPE& type of living quarters & type of living quarters is {\color{gray}\tt House, apartment, flat} \\
    LIVING\_OWNERSHIP& living quarters are owned, rented or occupied without payment & ownership status of living quarters is {\color{gray}\tt Owned or being bought by someone in the household} \\
    FOOD\_ASSISTANCE& received some form of food assistance & receipt of food assistance is {\color{gray}\tt No} \\
    WIC\_ASSISTANCE& percentage of year in which individual received assistance from WIC (Women, Infants, and Children supplemental program) & percentage of the year the respondent received WIC assistance is {\color{gray}\tt 0.00\%} \\
    SNAP\_ASSISTANCE& percentage of year in which individual received assistance from SNAP (Supplemental Nutrition Assistance Program) & percentage of the year the respondent received SNAP/food stamps assistance is {\color{gray}\tt 100.00\%} \\
    \multicolumn{3}{l}{\textit{health}} \\ 
    MEDICARE\_ASSISTANCE& percentage of year in which individual received assistance from MEDICARE & percentage of the year the respondent was covered by Medicare is {\color{gray}\tt 0.00\%} \\
    MEDICAID\_ASSISTANCE& percentage of year in which individual received assistance from MEDICAID & percentage of the year the respondent was covered by Medicaid is {\color{gray}\tt 0.00\%} \\
    HEALTHDISAB& has a physical, mental or other health condition limiting the amount of work they can do & work-limiting health condition status is {\color{gray}\tt Without a physical, mental or other health condition} \\
    DAYS\_SICK& number of days spent in bed due to illness or injury in the last year & number of days spent in bed due to illness or injury in the last year is {\color{gray}\tt 10 days} \\
    HOSPITAL\_NIGHTS& number of nights in hospital in the last year & number of nights in hospital last year is {\color{gray}\tt 0 nights} \\
    PRESCRIPTION\_MEDS& did take any prescription medications & use of prescription medication is {\color{gray}\tt Takes prescription medications} \\
    HEALTH\_INSURANCE\_PREMIUMS& expenditures for comprehensive health insurance premiums & amount paid for comprehensive health insurance premiums is {\color{gray}\tt \$0} \\
    HEALTH\_OVER\_THE\_COUNTER\_PRODUCTS\_PAY& out-of-pocket expenditures for over-the-counter health-related products & amount paid for over-the-counter health-related products is {\color{gray}\tt \$100} \\
    HEALTH\_MEDICAL\_CARE\_PAY& out-of-pocket expenditures on medical care & amount paid for non-premium medical out-of-pocket expenditures on medical care is {\color{gray}\tt \$1000} \\
    HEALTH\_HEARING& has serious difficulty hearing & hearing status is {\color{gray}\tt Deaf or with hearing difficulty} \\
    HEALTH\_SEEING& serious difficulty seeing & vision status is {\color{gray}\tt No vision difficulty} \\
    HEALTH\_COGNITIVE& has serious difficulty concentrating, remembering, or making decisions & cognition status is {\color{gray}\tt No cognitive difficulty} \\
    HEALTH\_AMBULATORY& has serious difficulty walking or climbing stairs & ambulatory status is {\color{gray}\tt No ambulatory difficulty} \\
    HEALTH\_SELF\_CARE& has difficulty dressing or bathing & self-care status is {\color{gray}\tt No self-care difficulty} \\
    HEALTH\_ERRANDS\_DIFFICULTY& has difficulty doing errands alone & independent-living status is {\color{gray}\tt No difficulty doing errands alone} \\
    HEALTH\_CORE\_DISABILITY& has at least one of six core disability measures & core disability status is {\color{gray}\tt With a core disability} \\
    HEALTH\_SUPPLEMENTAL\_DISABILITY& answered positively to at least one core questions, three child disability questions, or two work disability questions & supplemental disability status is {\color{gray}\tt With a supplemental disability} \\
    VISIT\_DOCTOR\_NUM& number of visits to a doctor, nurse, or any other type of medical provider & number of visits to a doctor, nurse, or any other type of medical provider is {\color{gray}\tt 2 visits} \\
    VISIT\_DENTIST\_NUM& number of visits to a dentist or other dental professional & number of dentist visits is {\color{gray}\tt 0 visits} \\
    \multicolumn{3}{l}{\textit{program participation details}} \\
    TANF\_ASSISTANCE& percentage of year in which individual received assistance from TANF & percentage of the year the respondent received TANF benefit is {\color{gray}\tt 0.00\%} \\
    TRANSPORTATION\_ASSISTANCE& received some kind of transportation assistance & receipt of transportation assistance is {\color{gray}\tt No} \\
    UNEMPLOYMENT\_COMP& received unemployment compensation payments at any time during the reference year & receipt of unemployment compensation payments is {\color{gray}\tt No} \\
    UNEMPLOYMENT\_COMP\_AMOUNT& total annual amount of unemployment compensation payments & amount of unemployment compensation in given year is {\color{gray}\tt \$0} \\
    SOCIAL\_SEC\_BENEFITS& received social security benefits for themselves or on behalf of a child & receipt of social security benefits is {\color{gray}\tt No} \\
    VA\_BENEFITS\_AMOUNT& total annual amount of VA benefits and pension & total amount of VA benefits and pension in given year is {\color{gray}\tt \$0} \\
    RETIREMENT\_INCOME\_AMOUNT& total annual amount of retirement income & total amount of retirement income in given year is {\color{gray}\tt \$0} \\
    SURVIVOR\_INCOME\_AMOUNT& total annual amount of survivor benefits & total amount of survivor benefits in given year is {\color{gray}\tt \$0} \\
    DISABILITY\_BENEFITS\_AMOUNT& total annual amount of payments due to sickness, accident or disability & total amount of disability benefits or income in given year is {\color{gray}\tt \$0} \\
    SEVERANCE\_PAY\_PENSION& received any severance pay or lump sum payments from a pension or retirement plan during the reference period & receipt of severance pay or a lump sum payment from a pension or retirement plan is {\color{gray}\tt No} \\
    FOSTER\_CHILD\_CARE\_AMT& amount of foster child care payments received in the reference year & amount of foster child care payments received in given year is {\color{gray}\tt \$0} \\
    CHILD\_SUPPORT\_AMT& amount of child support payments received in the reference year & amount of child support payments received in given year is {\color{gray}\tt \$0} \\
    ALIMONY\_AMT& amount of alimony payments received in the reference year & amount of alimony payments received in given year is {\color{gray}\tt \$0} \\
\end{longtable}
\end{center}
\vspace{-1em}
\footnotetext[1]{\url{https://www2.census.gov/programs-surveys/sipp/data/datasets/2014/}}
\end{landscape}

\FloatBarrier
\subsection{Models}\label{appendix:models}

The reported analyses use nine models that support matched comparisons with and without reasoning.

\begin{itemize}
    \item \textbf{Qwen (6 models total)}: Qwen 3 4B, 14B, and 32B~\citep{yang2025Qwen3TechnicalReport}, and Qwen 3.5 4B, 9B, and 27B~\citep{qwen2026Qwen35}, each with reasoning on/off.
    \item \textbf{OpenAI (3 models total)}: GPT 5.1~\citep{openai2025GPT51SystemCard}, GPT 5.2~\citep{openai2025GPT52SystemCard}, and GPT 5.4~\citep{openai2026GPT54SystemCard}, each at effort levels none/low/medium/high.
\end{itemize}

\FloatBarrier
\subsubsection{Prompting, Generation, and Response Processing}\label{appendix:inference-protocol}

\paragraph{Models and reasoning controls.}
Inference was run between July and September 2026. We use the public instruction-tuned Qwen checkpoints and access the OpenAI models through the API. For Qwen, we enable or disable reasoning through the model's chat template. For OpenAI, we request \texttt{none}, \texttt{low}, \texttt{medium}, or \texttt{high} reasoning effort. These controls are provider-specific; we do not assume that effort levels are comparable across developers.

\paragraph{Prompt templates.}
We use zero-shot numeric-QA prompts based on \texttt{folktexts}, with the same prompt for every model and reasoning setting. The system message is
\begin{quote}\small\ttfamily
You are a helpful assistant. You provide numeric probability estimates based on the information provided.
\end{quote}
followed by the answer-formatting instruction we add to \texttt{folktexts}:
\begin{quote}\small\ttfamily
Your response MUST end with your probability estimate in the following format:\\
Probability: X\%\\
where X is a number between 0 and 100.
\end{quote}
The user message describes the survey context, lists the individual's features in natural language, and asks the task-specific question:
\begin{quote}\small
\emph{ACSIncome:} What is the probability that this person's estimated yearly income is above \$50,000 ?\\
\emph{ACSEmployment:} What is the probability that this person is an employed civilian?\\
\emph{SIPP:} What is the probability that the income of this individual's household is at least three times the poverty threshold based on the Official Poverty Measure (OPM)?
\end{quote}
Feature mappings are given in Appendix~\ref{appendix:tasks}. We add no chain-of-thought instruction: reasoning is controlled through the model settings, while prompt wording is held fixed.

\paragraph{Generation settings.}
We use temperature $1.0$ in both reasoning conditions, following the requirements or recommendations of the evaluated models. Other sampling parameters retain their provider or backend defaults (for vLLM, top-$p=1$ with no top-$k$ filtering). The initial generation budget is $8000$ tokens per response; input prompts are not truncated. 

\paragraph{Repeated evaluation.}
We repeat each model-reasoning-setting configuration with different generation seeds while holding the evaluation data and prompts fixed. Qwen \texttt{on}/\texttt{off} and OpenAI \texttt{none}/\texttt{high} use five resamples; OpenAI \texttt{low}/\texttt{medium} use three. These repeats capture variability in generated responses, rather than variability from drawing new evaluation samples.

\paragraph{Reasoning-length measurement.}
For OpenAI, we use the API-reported reasoning-token count, rather than the length of the returned reasoning summary. For Qwen, we count tokens in the delimited reasoning portion of the response, including its boundary markers and excluding the final answer. Responses whose reasoning blocks remained unclosed at the initial token limit were regenerated with a $16{,}000$-token budget. The remaining unclosed responses (less than $0.2\%$ in numeric-QA mode) were assigned a reasoning-token count of zero. Counts therefore follow each provider's available representation of reasoning; relative-length analyses use within-model ranks or standardization.

\paragraph{Answer extraction.}
We extract probabilities from the final answer, excluding the reasoning trace. The extraction heuristic prioritizes the requested \texttt{Probability: X\%} format, with fallbacks for other percentage or decimal expressions. When the selected format occurs more than once, we use its last occurrence to allow for revised estimates. We convert percentages to probabilities and reject values outside $[0,1]$. Unparsable answers are treated as missing rather than assigned a default probability.

\paragraph{Missing responses.}
When no valid answer was returned because of truncation, a request failure, or an unparsable response, we sampled another response for the affected individual. This yielded valid answers for nearly all evaluated inputs. Analyses use only individuals with valid answers across all compared models and reasoning settings, so comparisons are made on the same individuals.

\FloatBarrier
\subsection{General Performance and Risk Scores}\label{appendix:general-performance}
We assess accuracy comparability using an \emph{empirical Rashomon set}~\citep{breiman2001StatisticalModelingTwo,fisher2019AllModelsAre}: the evaluated configurations whose accuracy is within $\epsilon=0.05$ (five percentage points) of the best evaluated configuration on each task. We retain configurations outside this set to preserve matched comparisons across reasoning settings.

\begin{figure}[htbp]
    \centering
    \includegraphics[]{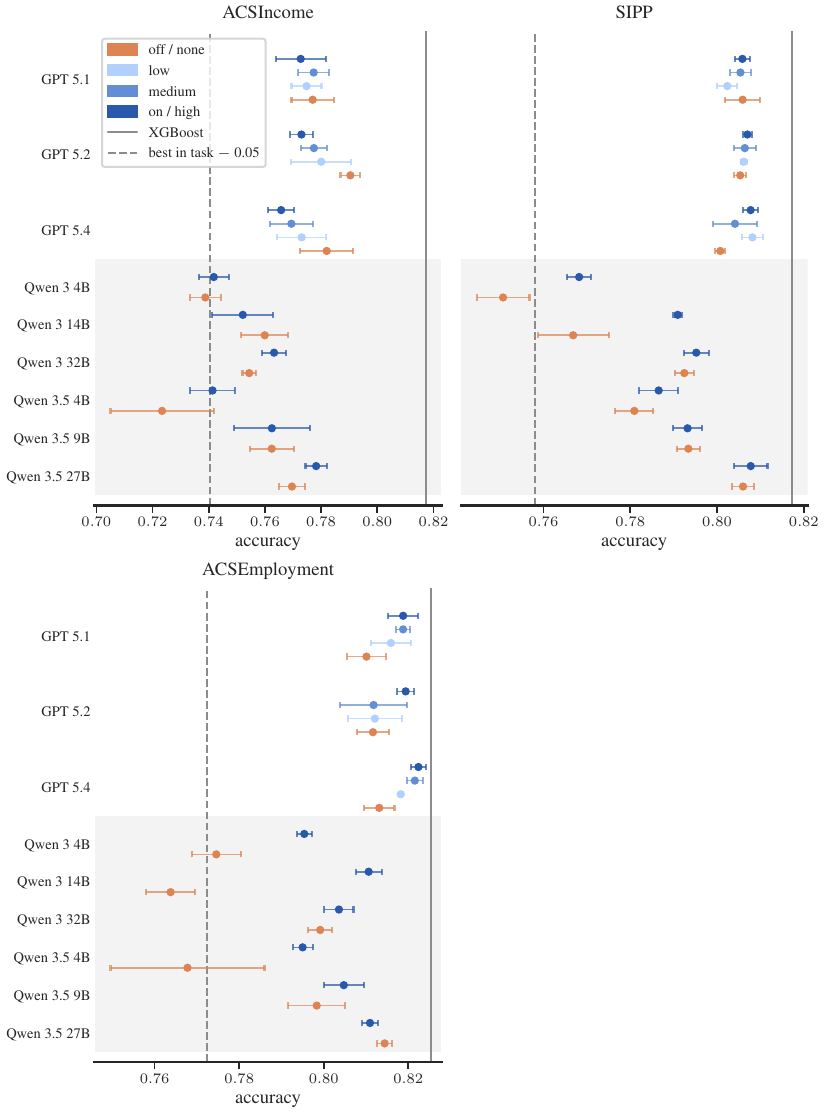}
    \caption{Accuracy of each model and reasoning setting per task. Dots show the mean over resamples and error bars the spread across resamples; the solid line is an XGBoost baseline and the dashed line the Rashomon-set cutoff (best configuration minus $\epsilon=0.05$).}
    \label{fig:appendix-general-accuracy-all-tasks}
\end{figure}
Accuracy varies considerably across models (\autoref{fig:appendix-general-accuracy-all-tasks}). No configuration reaches the XGBoost baseline; OpenAI models come closest, in particular on ACSEmployment. Enabling reasoning improves accuracy for most Alibaba models, whereas for OpenAI models the differences between effort levels are small and not consistent in direction. Most configurations lie within $\epsilon=0.05$ of the best configuration in each task; the exceptions, based on the plotted means, are Qwen 3 4B and Qwen 3.5 4B on ACSIncome, Qwen 3 4B on SIPP, and Qwen 3 14B and Qwen 3.5 4B on ACSEmployment, all with reasoning disabled.

\begin{figure}[htbp]
    \centering
    \begin{subfigure}[b]{\textwidth}
        \centering
        \includegraphics[width=\textwidth,height=0.26\textheight,keepaspectratio]{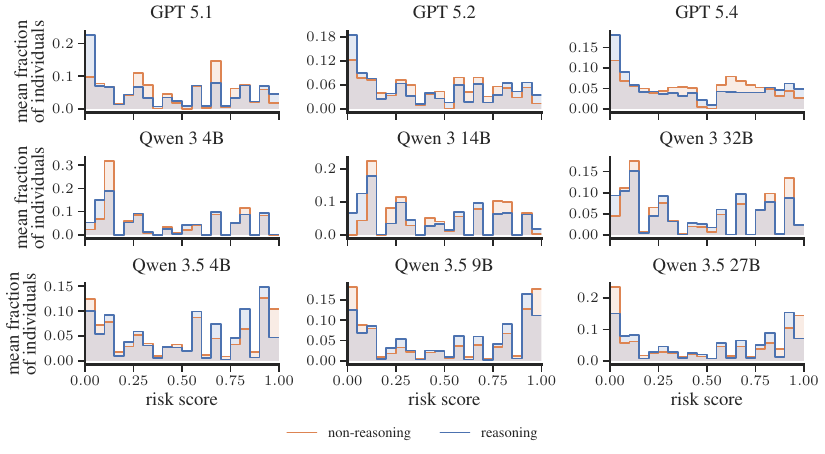}
        \caption{ACSIncome}
        \label{fig:appendix-general-risk-scores-ACSIncome}
    \end{subfigure}
    \hfill
    \begin{subfigure}[b]{\textwidth}
        \centering
        \includegraphics[width=\textwidth,height=0.26\textheight,keepaspectratio]{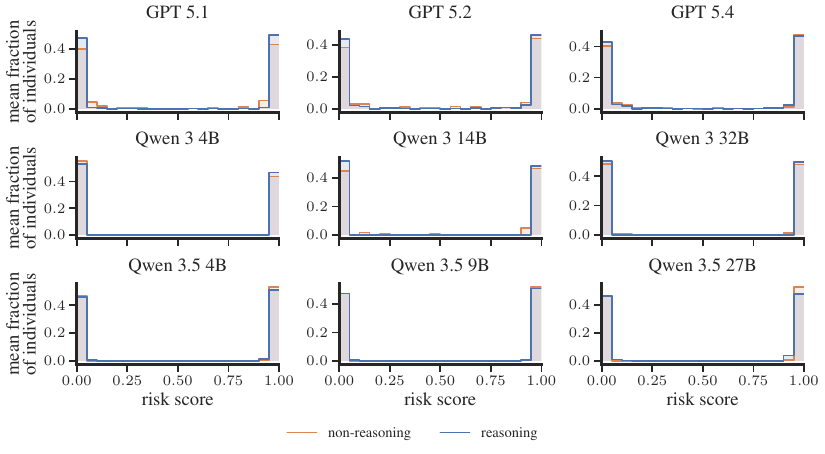}
        \caption{SIPP}
        \label{fig:appendix-general-risk-scores-SIPP}
    \end{subfigure}
    \hfill
    \begin{subfigure}[b]{\textwidth}
        \centering
        \includegraphics[width=\textwidth,height=0.26\textheight,keepaspectratio]{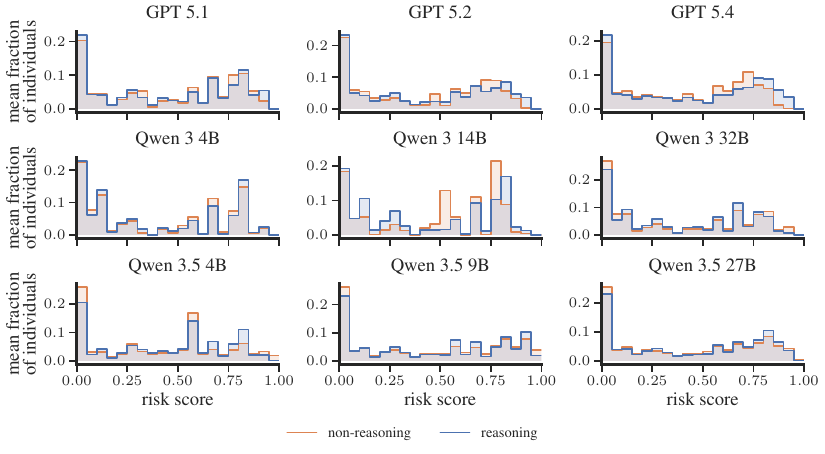}
        \caption{ACSEmployment}
        \label{fig:appendix-general-risk-scores-ACSEmployment}
    \end{subfigure}
    \caption{Distribution of risk scores per model with reasoning disabled (orange) and enabled (blue), pooled over resamples.}
    \label{fig:appendix-general-risk-scores-all-tasks}
\end{figure}
Risk-score distributions differ more across tasks than between reasoning settings (\autoref{fig:appendix-general-risk-scores-all-tasks}). On SIPP, scores concentrate near $0$ and $1$ in both conditions; on the ACS tasks, models use a wider range of scores, often concentrated at a few values. Reasoning changes these distributions without a uniform shift toward the extremes. The plots therefore do not suggest a general increase in score extremity that would by itself explain the higher agreement.

\FloatBarrier
\subsection{Reasoning as Intervention: Additional Results}
These analyses examine how enabling reasoning changes shared predictions. We first separate observed agreement from its accuracy-based baseline, then distinguish shared correct predictions from shared errors and examine the resulting changes in individual recourse.

\FloatBarrier
\subsubsection{\texorpdfstring{$\kappa$}{kappa} for Varying Reasoning Levels}
\begin{figure}[htbp]
    \centering
    \includegraphics[width=\linewidth]{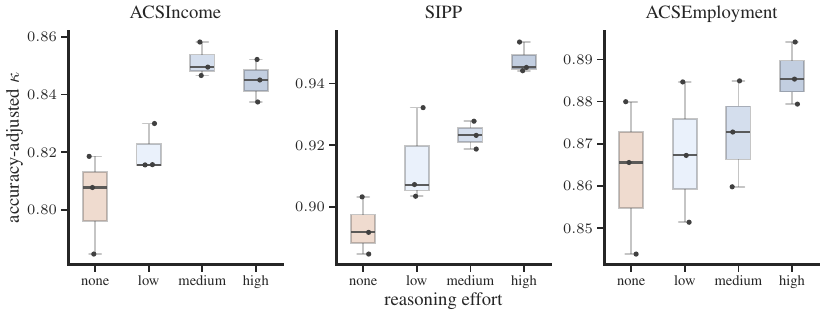}
    \caption{Accuracy-adjusted $\kappa$ between pairs of different OpenAI models at each reasoning effort level.}
    \label{fig:appendix-effort-kappa-by-level-all-tasks}
\end{figure}
\autoref{fig:appendix-effort-kappa-by-level-all-tasks} examines whether agreement changes gradually with requested reasoning effort. Among the three {OpenAI} model pairs, agreement generally rises from \texttt{none} to \texttt{high} on SIPP and ACSEmployment. On ACSIncome, it peaks at \texttt{medium} before declining slightly. Thus, enabling reasoning and increasing it further need not have identical effects, and the pattern varies slightlyby task.

\FloatBarrier
\subsubsection{\texorpdfstring{$\kappa$}{kappa} Components}\label{appendix:effort-kappa-components}
\begin{figure}[htbp]
    \centering
    \includegraphics[width=\linewidth]{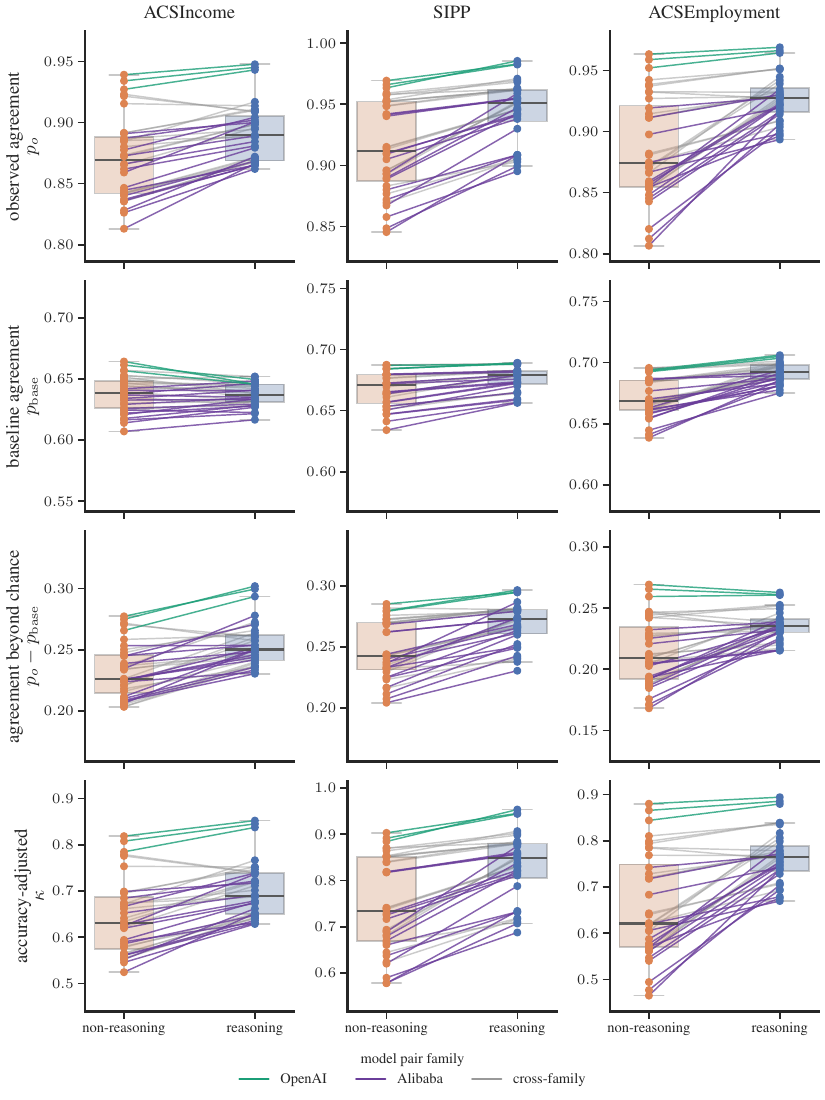}
    \caption{Components of accuracy-adjusted $\kappa$ with reasoning disabled (orange) and enabled (blue): observed agreement $p_o$, baseline agreement $p_\mathrm{base}$, their difference, and $\kappa$. Lines connect the same model pair (green: OpenAI, purple: Alibaba, gray: cross-developer). Pairs are formed only between configurations with the same reasoning condition (both disabled or both enabled).}
    \label{fig:appendix-effort-kappa-components-all-tasks}
\end{figure}
\autoref{fig:appendix-effort-kappa-components-all-tasks} separates changes in observed agreement from changes expected under the accuracy-based baseline. Reasoning raises observed agreement for most model pairs, while the baseline shifts less and in a task-dependent direction. Agreement beyond the baseline and the normalized measure $\kappa$ generally increase, particularly for Alibaba pairs. The higher $\kappa$ therefore reflects changes in joint predictions beyond those accounted for by individual model accuracies.

\FloatBarrier
\subsubsection{Pairwise Change in Agreement}
\begin{figure}[htbp]
    \centering
    \includegraphics[width=\linewidth]{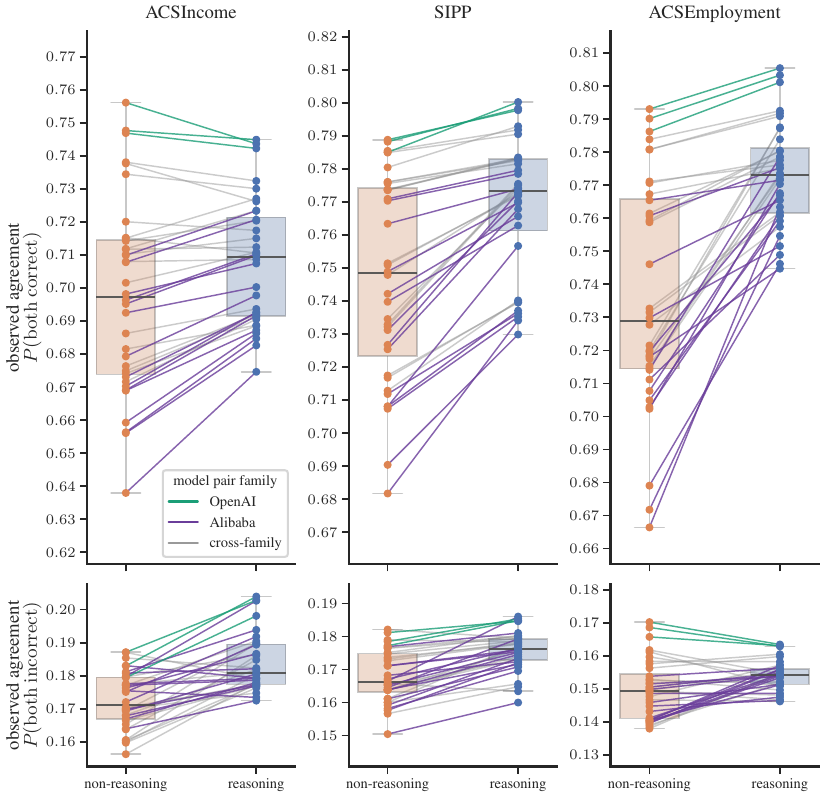}
    \caption{Observed pairwise agreement on correct predictions (\textit{top}) and on incorrect predictions (\textit{bottom}), with reasoning disabled (orange) and enabled (blue). Dots give the share of individuals for whom both models are correct (top) or both are incorrect (bottom). Lines connect the same model pair (green: OpenAI; purple: Alibaba; gray: cross-developer). Axis limits vary across panels, so visual slopes are not directly comparable. Pairs are formed only between configurations with the same reasoning condition (both disabled or both enabled).}
    \label{fig:appendix-effort-pairwise-agreement-by-correctness-all-tasks}
\end{figure}
\autoref{fig:appendix-effort-pairwise-agreement-by-correctness-all-tasks} separates agreement into both models being correct and both being incorrect. For Alibaba pairs, reasoning generally increases both components. OpenAI pairs show more task-dependent changes: on ACSIncome, agreement shifts toward shared errors, whereas on ACSEmployment shared correct predictions increase and shared errors decrease. Higher aggregate agreement can therefore accompany different changes in prediction quality.

\FloatBarrier
\subsubsection{Recourse Levels}\label{appendix:effort-recourse-distribution}
\begin{figure}[htbp]
    \centering
    \includegraphics[width=\linewidth]{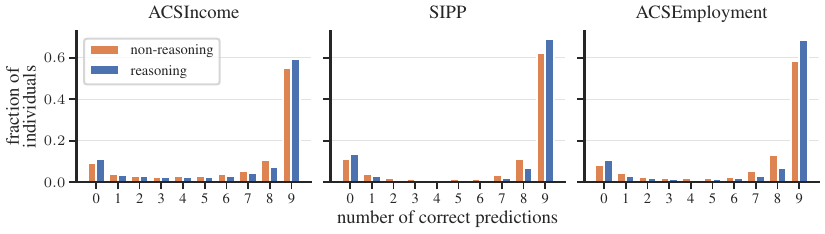}
    \caption{Distribution of recourse levels with reasoning disabled (orange) and enabled (blue): share of individuals by the number of models (out of $M=9$) that predict their label correctly, from $0$ (no recourse) to $9$ (full recourse). Each distribution counts the nine models within one reasoning condition (all disabled or all enabled).}
    \label{fig:appendix-effort-recourse-distribution}
\end{figure}

\autoref{fig:appendix-effort-recourse-distribution} shows how many of the nine models predict each individual correctly. Most individuals receive a correct prediction from every model, and reasoning increases this share on all three tasks. It also increases the share for whom every model is wrong, while reducing the mass at seven or eight correct models. This concentration at both extremes explains why greater agreement can coexist with less opportunity to obtain a correct prediction by switching models.

\FloatBarrier
\subsubsection{Recourse Levels by Label}

\begin{figure}[htbp]
    \centering
    \includegraphics[width=\linewidth]{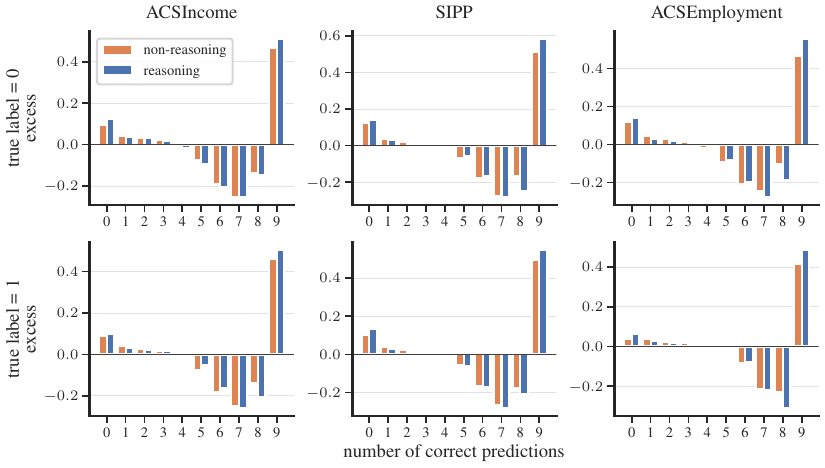}
    \caption{Excess over the independent-error baseline for each recourse level, conditioned on the true label (\textit{top:} label $0$, \textit{bottom:} label $1$), with reasoning disabled (orange) and enabled (blue). Each distribution counts the nine models within one reasoning condition (all disabled or all enabled).}
    \label{fig:appendix-effort-recourse-by-label-all-tasks}
\end{figure}
Splitting the population by true label preserves the concentration at extreme recourse levels (\autoref{fig:appendix-effort-recourse-by-label-all-tasks}). For both labels, reasoning increases the excess over the plotted baseline at full recourse and, more modestly, at no recourse. Levels with most but not all models correct generally remain below the baseline. The aggregate pattern is therefore present within both outcome groups, although its magnitude differs across tasks and labels.

\FloatBarrier
\subsubsection{Recourse Level Transitions}

\begin{figure}[htbp]
    \centering
    \includegraphics[width=\linewidth]{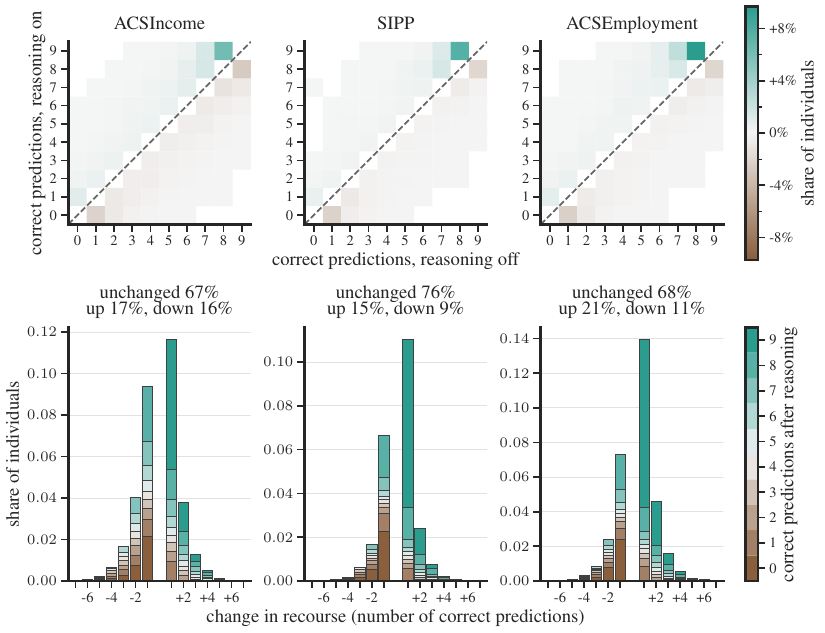}
    \caption{Change in recourse levels when reasoning is enabled. \textit{Top:} transitions from the number of correct models with reasoning disabled ($x$-axis) to enabled ($y$-axis). Color magnitude gives the share of individuals making each transition; increases are positive (green), decreases negative (brown), and unchanged cases on the diagonal are omitted. \textit{Bottom:} share of all individuals by the change in the number of correct models, excluding unchanged cases and stacked by the number correct with reasoning enabled.}
    \label{fig:appendix-effort-recourse-transitions-all-tasks}
\end{figure}
Most individuals retain the same recourse level when reasoning is enabled: $67\%$ on ACSIncome, $76\%$ on SIPP, and $68\%$ on ACSEmployment (\autoref{fig:appendix-effort-recourse-transitions-all-tasks}). Increases are more common than decreases, and changes are usually by one or two correct models. The prominent transition from eight to nine correct models contributes to higher full recourse, while transitions from one correct model to none show how some individuals lose their only correct prediction.

\FloatBarrier
\subsection{Reasoning as Behavior: Additional Results}\label{appendix:effort-vs-length}
These analyses relate realized reasoning length to agreement and recourse within fixed reasoning settings. We check whether the patterns persist across model families, model pairs, and alternative length measures. These comparisons describe associations; they do not isolate the effect of making a model reason longer.

\FloatBarrier
\subsubsection{Reasoning Length by Effort}

\begin{figure}[htbp]
    \centering
    \begin{subfigure}[b]{\textwidth}
        \centering
        \includegraphics[width=\textwidth,height=0.26\textheight,keepaspectratio]{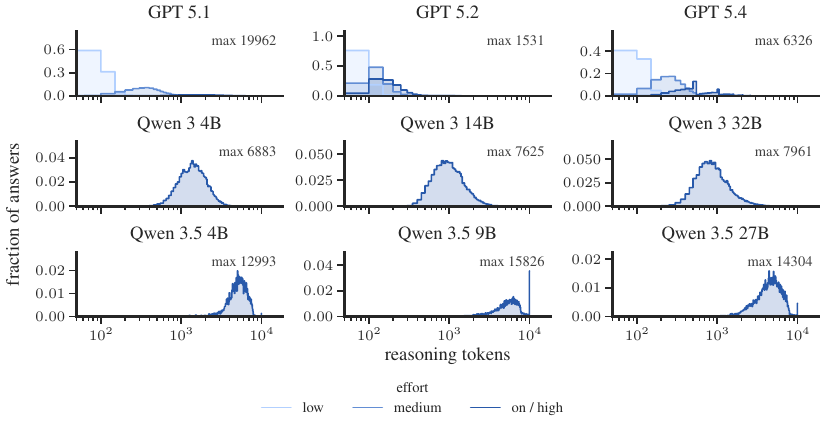}
        \caption{ACSIncome}
        \label{fig:appendix-length-token-distributions-by-effort-ACSIncome}
    \end{subfigure}
    \hfill
    \begin{subfigure}[b]{\textwidth}
        \centering
        \includegraphics[width=\textwidth,height=0.26\textheight,keepaspectratio]{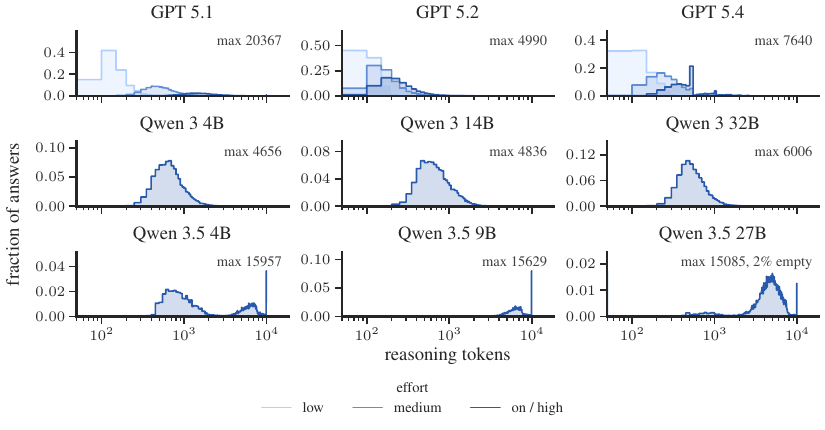}
        \caption{SIPP}
        \label{fig:appendix-length-token-distributions-by-effort-SIPP}
    \end{subfigure}
    \hfill
    \begin{subfigure}[b]{\textwidth}
        \centering
        \includegraphics[width=\textwidth,height=0.26\textheight,keepaspectratio]{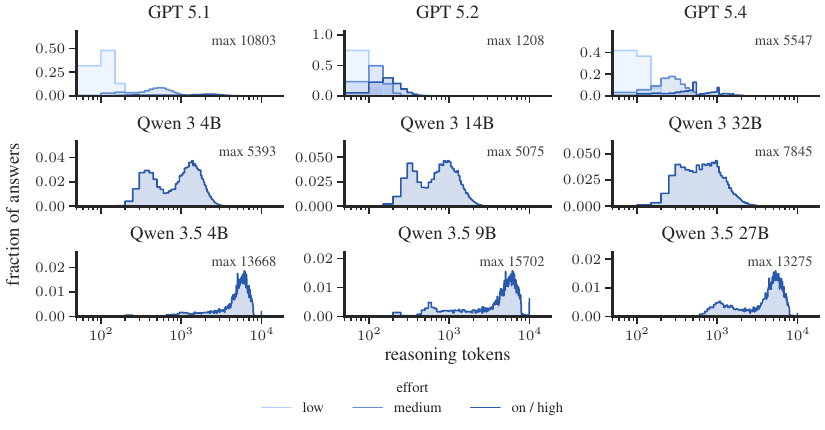}
        \caption{ACSEmployment}
        \label{fig:appendix-length-token-distributions-by-effort-ACSEmployment}
    \end{subfigure}
    \caption{Distribution of reasoning tokens per answer for each model and enabled reasoning effort (logarithmic $x$-axis). The $y$-axis gives the fraction of answers in each token bin; colors denote OpenAI \texttt{low}, \texttt{medium}, and \texttt{high} effort or Alibaba \texttt{on}. Numbers give the largest observed length for each model and task.}
    \label{fig:appendix-length-token-distributions-by-effort-all-tasks}
\end{figure}
\autoref{fig:appendix-length-token-distributions-by-effort-all-tasks} shows that requested effort and realized reasoning length are related but distinct. Higher OpenAI effort settings generally produce longer traces, with substantial overlap between settings. Lengths also vary markedly across models: Qwen 3.5 often generates several thousand tokens, while many OpenAI responses are much shorter. These differences motivate comparing relative length within models rather than treating a fixed token count as equivalent across models.

\FloatBarrier
\subsubsection{Agreement and Reasoning Length by Developer}

\begin{figure}[htbp]
    \centering
    \begin{subfigure}[b]{\textwidth}
        \centering
         \includegraphics[width=\linewidth]{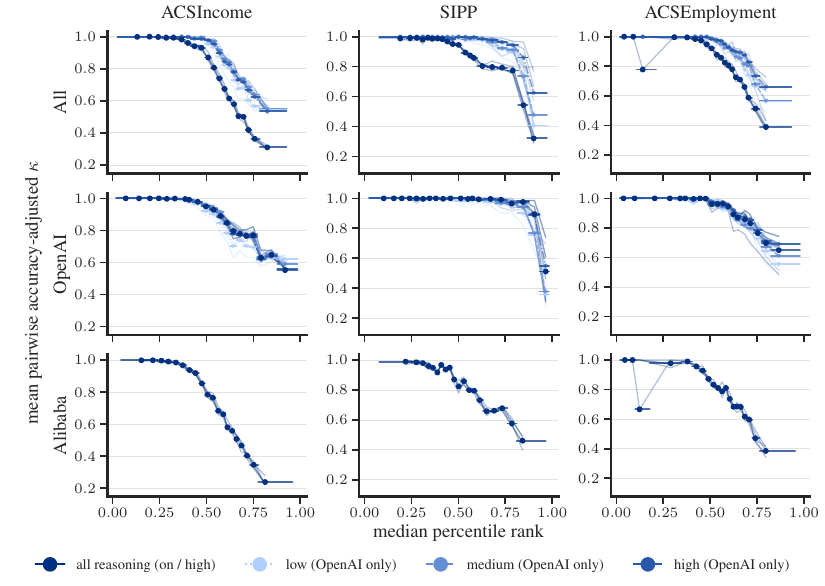}
        \caption{Per developer.}
        \label{fig:appendix-length-agree-by-dev}
    \end{subfigure}
    \hfill
    \begin{subfigure}[b]{\textwidth}
        \centering
        \includegraphics[width=\linewidth]{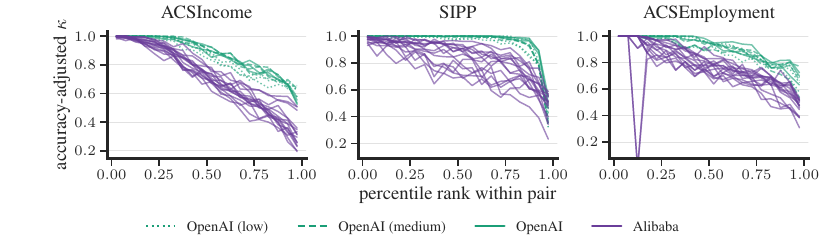}
        \caption{Per model pair.}
        \label{fig:appendix-length-agree-pairs-by-dev}
    \end{subfigure}
    \caption{Accuracy-adjusted pairwise agreement $\kappa$ by relative reasoning length, with percentile-rank bins formed (a) separately per developer and (b) separately per model pair. Rows in (a) show all models, OpenAI models, and Alibaba models. Each line in (b) represents one model pair at one reasoning setting (green: OpenAI; purple: Alibaba); dotted, dashed, and solid green lines denote \texttt{low}, \texttt{medium}, and \texttt{high} effort, respectively. Bins need not contain the same individuals across developers in (a) or across pairs in (b); the panels compare trends rather than identical groups of individuals.}
    \label{fig:appendix-length-agree-by-dev-details}
\end{figure}

\begin{figure}[htbp]
    \centering
    \begin{subfigure}[b]{\textwidth}
        \centering
         \includegraphics[width=\linewidth,height=0.36\textheight,keepaspectratio]{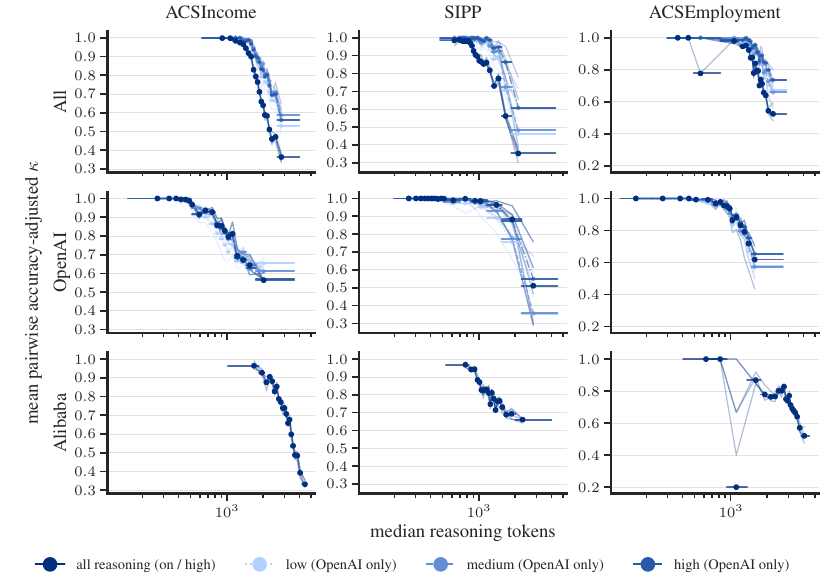}
        \caption{Raw reasoning length.}
        \label{fig:appendix-length-agree-by-dev-raw}
    \end{subfigure}
    \hfill
    \begin{subfigure}[b]{\textwidth}
        \centering
        \includegraphics[width=\textwidth,height=0.36\textheight,keepaspectratio]{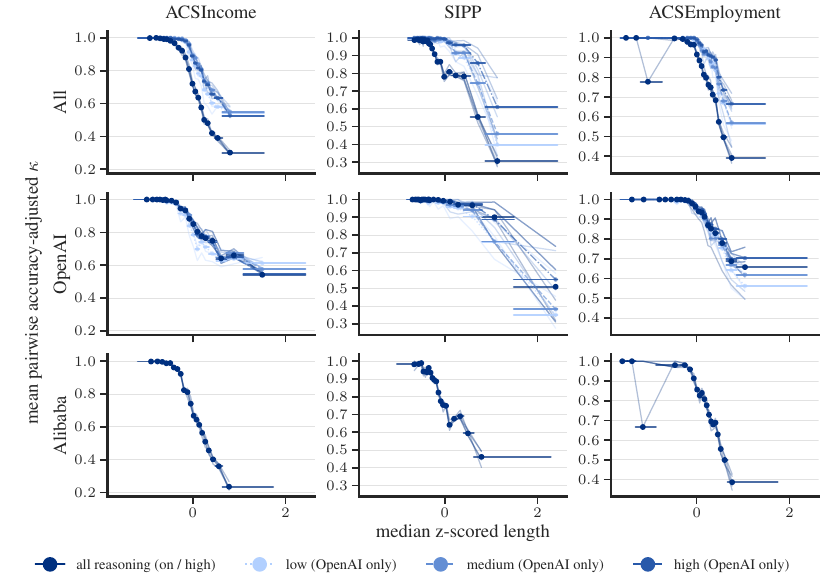}
        \caption{Z-scored reasoning length.}
        \label{fig:appendix-length-agree-by-dev-zscore}
    \end{subfigure}
    \caption{Accuracy-adjusted pairwise agreement $\kappa$ by reasoning length, measured (a) in raw tokens (logarithmic $x$-axis) and (b) as z-scores within each model. Rows show all models, OpenAI models, and Alibaba models, with bins formed separately using the models in each row. Dark blue denotes reasoning enabled (Alibaba \texttt{on}, OpenAI \texttt{high}); lighter curves show OpenAI-only comparisons at \texttt{low}, \texttt{medium}, and \texttt{high} effort.}
    \label{fig:appendix-length-agree-by-dev-x-axis}
\end{figure}

\paragraph{Binning by developer.}
\autoref{fig:appendix-length-agree-by-dev} tests whether the length--agreement association persists when each developer defines its own length bins. Agreement generally decreases toward longer traces for both developers across the three tasks. This shows that the overall trend is also visible within each model family. Because bins are constructed separately, comparisons across rows concern the shape of the relationship rather than the same groups of individuals.

\paragraph{Binning by model pair.}
\autoref{fig:appendix-length-agree-pairs-by-dev} goes further by constructing length bins separately for each model pair. Every pair shows lower agreement at longer relative lengths on all three tasks, for Alibaba pairs and for OpenAI pairs at every effort level. The association therefore holds across pairs, although each curve refers to its own groups of individuals.

\paragraph{Measuring reasoning length.}
\autoref{fig:appendix-length-agree-by-dev-x-axis} checks whether the result depends on measuring length by percentile rank. The decline generally remains when using raw token counts or within-model z-scores. Absolute token counts differ substantially across models, so relative measures facilitate comparisons while preserving many of the observed trends.

\FloatBarrier
\subsubsection{Observed Agreement and Reasoning Length}
\begin{figure}[htbp]
    \centering
    \includegraphics[width=\linewidth]{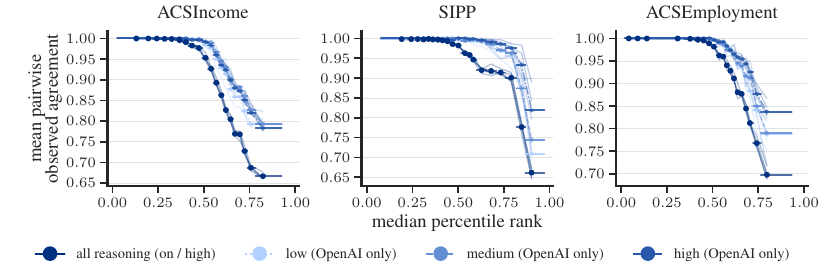}
    \caption{Mean pairwise observed agreement $p_o$ by reasoning-length bin, using the same binning as in \autoref{fig:agree-length}: individuals are grouped by their median reasoning-length percentile rank across the nine models with reasoning enabled. Dark blue shows all nine models with reasoning enabled (Alibaba \texttt{on}, OpenAI \texttt{high}); lighter curves show the three OpenAI models at each indicated effort level. Unlike $\kappa$, $p_o$ is not adjusted for agreement expected from model accuracy.}
    \label{fig:appendix-length-agree-observed}
\end{figure}
Observed agreement also falls toward longer reasoning traces (\autoref{fig:appendix-length-agree-observed}), reaching approximately $0.66$--$0.70$ in the longest bins across all nine models. The decline is therefore visible before adjusting for agreement expected from accuracy. OpenAI-only comparisons show a similar tail decline, with differences across effort levels, especially on SIPP. This supports interpreting the main result as a change in shared predictions rather than solely a consequence of the normalization in $\kappa$.

\FloatBarrier
\subsubsection{Recourse Levels and Reasoning Length}
\begin{figure}[htbp]
    \centering
    \includegraphics[width=\linewidth]{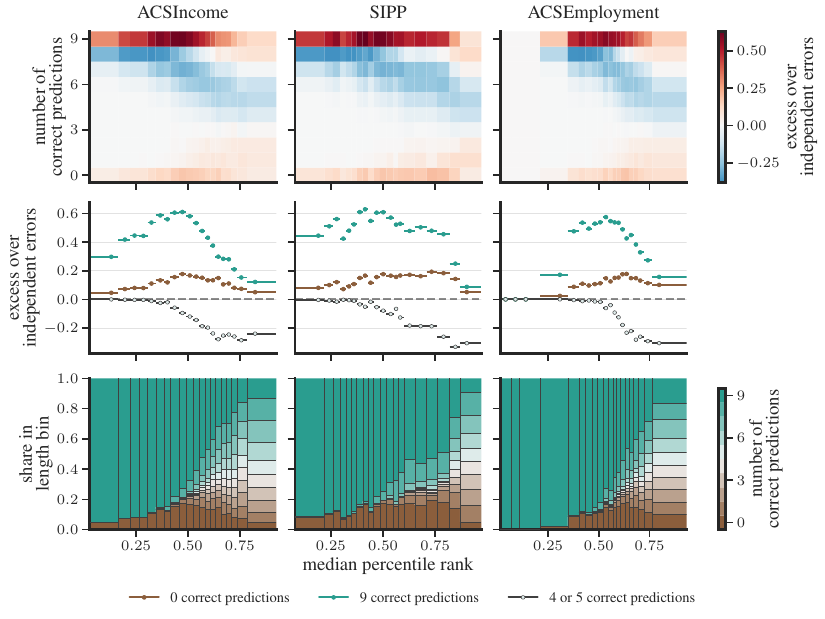}
    \caption{Recourse levels by reasoning-length bin across all nine models with reasoning enabled (Alibaba \texttt{on}, OpenAI \texttt{high}), using the binning of \autoref{fig:agree-length}. Recourse level $m/9$ is the fraction of models predicting an individual correctly. \textit{Top:} excess over the independent-error baseline for each number $m$ of correct models. \textit{Middle:} excess at no recourse ($m=0$), full recourse ($m=9$), and the maximal split ($m\in\{4,5\}$). \textit{Bottom:} observed distribution of recourse levels. Baselines use accuracies within each bin; distributions and baselines are computed per resample and averaged over five resamples.}
    \label{fig:appendix-length-vote-excess-joint}
\end{figure}
\autoref{fig:appendix-length-vote-excess-joint} connects the pairwise results to individual recourse. Short traces are dominated by unanimous predictions, whereas longer traces more often accompany a mixture of correct and incorrect models. Relative to the independent-error baseline, full recourse has its largest excess around intermediate lengths and then declines; no recourse remains overrepresented in bins where models make errors. Greater disagreement at long lengths thus does not eliminate shared failure.

\begin{figure}[htbp]
    \centering
    \includegraphics[width=\linewidth]{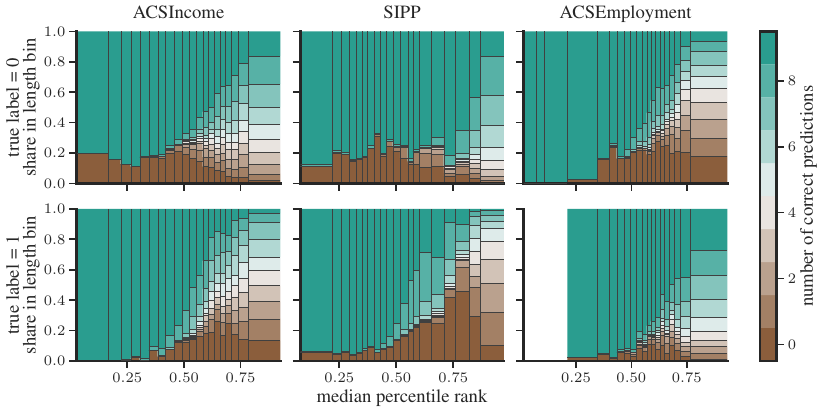}
    \caption{Observed distribution of recourse levels by reasoning-length bin, conditioned on the true label (\textit{top:} $y=0$; \textit{bottom:} $y=1$). Recourse level $m/9$ is the fraction of the nine models predicting an individual correctly, with reasoning enabled (Alibaba \texttt{on}, OpenAI \texttt{high}). Colors indicate the number $m$ of correct models, and each bar gives shares within a length bin and label group. The $x$-axis is the median reasoning-length percentile rank, as in \autoref{fig:agree-length}.}
    \label{fig:appendix-length-recourse-by-label}
\end{figure}
\autoref{fig:appendix-length-recourse-by-label} shows the corresponding distributions separately for each true label. Short traces are mostly associated with unanimous predictions in both groups, while intermediate recourse levels become more common toward longer traces. The composition of unanimous outcomes differs by label: for positive instances on ACSIncome and SIPP, shared errors are more prevalent in parts of the long-reasoning range. Empty short-length bins for positive ACSEmployment instances reflect the absence of those individuals in those bins.

\end{document}